\documentclass[11pt,a4paper,twocolumn]{article}

\usepackage{fontspec}
\newfontfamily\devanagarifont[Path=./]{NotoSansDevanagari-Regular.ttf}[Script=Devanagari, Scale=0.95]
\newcommand{\hi}[1]{{\devanagarifont #1}}
\usepackage{latexsym}
\usepackage[final]{microtype}        

\usepackage[margin=2cm,columnsep=0.5cm]{geometry}
\usepackage{amsmath}
\usepackage{amssymb}
\usepackage{amsfonts}

\usepackage{booktabs}
\usepackage{multirow}
\usepackage{array}
\usepackage{tabularx}
\usepackage{makecell}

\usepackage{graphicx}
\usepackage{float}
\usepackage[font=small,labelfont=bf]{caption}
\usepackage{subcaption}

\usepackage[dvipsnames]{xcolor}
\usepackage[breaklinks,colorlinks,citecolor=MidnightBlue,linkcolor=BrickRed,urlcolor=MidnightBlue]{hyperref}

\usepackage{natbib}
\usepackage{enumitem}
\usepackage{algorithm}
\usepackage{algorithmic}
\usepackage{footnote}

\usepackage{fancyhdr}
\fancypagestyle{plain}{%
  \fancyhf{}%
}

\newcommand{\ours}{\textsc{XKD-Dial}}
\newcommand{\flantbase}{Flan-T5-Base}
\newcommand{\flantlarge}{Flan-T5-Large}
\newcommand{\flantxl}{Flan-T5-XL}
\newcommand{\mistral}{Mistral-7B}
\newcommand{\llama}{LLaMA-3.2-1B}
\newcommand{\gemma}{Gemma-2-2B}

\newcommand{\tup}{\,{\scriptsize$\uparrow$}}
\newcommand{\tdn}{\,{\scriptsize$\downarrow$}}
\newcommand{\tsm}{\,{\scriptsize$\sim$}}

\title{
\textbf{Progressive Training for Explainable Citation-Grounded Dialogue: \\
Reducing Hallucination to Zero in English-Hindi LLMs}
}

\author{
\textbf{Vedant Pandya} \\
School of Artificial Intelligence and Data Engineering (SAIDE) \\
Indian Institute of Technology Jodhpur \\
\texttt{m25ai1132@iitj.ac.in}
}

\date{}

\begin{document}
\maketitle

\begin{abstract}
Knowledge-grounded dialogue systems aim to generate informative, contextually relevant responses by conditioning on external knowledge sources. However, most existing approaches focus exclusively on English, lack explicit citation mechanisms for verifying factual claims, and offer limited transparency into model decision-making. We present \ours{}, a progressive four-stage training pipeline for \textbf{explainable, knowledge-grounded dialogue generation} in a bilingual (English--Hindi) setting, comprising: (1)~multilingual adaptation, (2)~English dialogue SFT with citation grounding, (3)~bilingual dialogue SFT, and (4)~GRPO alignment with citation-aware rewards. We evaluate six models spanning encoder-decoder (250M--3B) and decoder-only (1B--7B) architectures at every pipeline stage. Our key contributions are: (i)~three post-hoc explainability analyses - cross-attention alignment, Integrated Gradients attribution, and occlusion-based causal grounding - applied systematically across the training trajectory to reveal \emph{how} citation behaviour is learned, not only \emph{whether} it is learned; (ii)~citation-grounded SFT reduces hallucination to 0.0\% for encoder-decoder models from Stage~2 onward; (iii)~the progressive pipeline prevents catastrophic forgetting while improving Hindi capabilities; (iv)~smaller models match larger models on English after SFT; and (v)~GRPO provides marginal improvement over well-designed SFT for structured citation tasks. We evaluate across six automatic metrics (BLEU, ROUGE, BERTScore, FactScore, Citation-F1, hallucination rate).

\end{abstract}


\section{Introduction}
\label{sec:introduction}

Knowledge-grounded dialogue generation has emerged as a critical research direction for building conversational AI systems that produce factually accurate, informative responses~\citep{dinan2019wizard, rashkin2021faithdial}. By conditioning response generation on retrieved knowledge passages, these systems can mitigate the chronic hallucination problem of large language models (LLMs) - where models generate plausible-sounding but factually incorrect information~\citep{ji2023survey}. However, current approaches suffer from three fundamental limitations.

\textbf{First, the monolingual bottleneck.} The vast majority of knowledge-grounded dialogue research focuses exclusively on English~\citep{dinan2019wizard, rashkin2021faithdial, kim2020dstc9}. For languages like Hindi - spoken by over 600 million people - there exists no standard benchmark, no established training methodology, and no systematic study of how knowledge-grounded dialogue systems perform in a bilingual setting. Extending such systems to Hindi is particularly challenging due to: (a)~limited availability of Hindi dialogue corpora with knowledge annotations, (b)~morphological richness and free word order that complicate both generation and evaluation, and (c)~the need to handle code-switching and cross-lingual knowledge transfer.

\textbf{Second, the absence of verifiable citations.} While retrieval-augmented generation (RAG) systems retrieve relevant passages~\citep{lewis2020rag, guu2020realm}, the generated response typically does not indicate \emph{which} passage supports \emph{which} claim. Without explicit citation markers (e.g., ``According to [1], \ldots''), users cannot verify factual claims against their sources, undermining trust and transparency. Recent work on attributed text generation~\citep{rashkin2021faithdial} has highlighted this gap, but citation-grounded training for dialogue remains underexplored.

\textbf{Third, the opacity of model decisions.} Even when a model generates a correct, grounded response, it provides no insight into \emph{why} it selected particular knowledge passages or \emph{how} it composed the response. This opacity is especially problematic for citation-grounded systems: a model may produce the correct citation marker \texttt{[1]} without genuinely conditioning its output on passage~1, making citation accuracy an unreliable quality signal on its own. Interpretability methods - cross-attention visualization~\citep{jain2019attention, wiegreffe2019attention}, Integrated Gradients~\citep{sundararajan2017axiomatic}, and occlusion-based causal grounding~\citep{lei2016rationalizing} - can expose this dissociation, but their systematic application to knowledge-grounded dialogue generation across an entire training trajectory has not been attempted.

\subsection{Our Approach}

We propose \ours{} (\textbf{E}xplainable \textbf{K}nowledge-Grounded \textbf{D}ialogue), a progressive four-stage training pipeline designed to address all three limitations simultaneously. Our key insight is that complex multilingual, knowledge-grounded generation capabilities can be built incrementally, where each training stage adds a specific skill while preserving previously learned capabilities:

\begin{enumerate}[leftmargin=*, itemsep=2pt]
    \item \textbf{Stage~1: Multilingual Adaptation.} English--Hindi translation training to build bilingual representations, particularly for models with limited Hindi pretraining exposure.
    \item \textbf{Stage~2: English Dialogue SFT.} Supervised fine-tuning on English knowledge-grounded dialogue with explicit citation markers, teaching the model to generate responses that attribute claims to specific knowledge passages.
    \item \textbf{Stage~3: Bilingual Dialogue SFT.} Extension to Hindi dialogue with citations, leveraging cross-lingual transfer from Stage~2.
    \item \textbf{Stage~4: GRPO Alignment.} Reinforcement learning via Group Relative Policy Optimization~\citep{shao2024deepseekmath} with a composite reward function that incentivizes citation accuracy, factual consistency, and penalizes hallucination.
\end{enumerate}

\subsection{Contributions}

Our main contributions are as follows:

\begin{enumerate}[leftmargin=*, itemsep=2pt]
    \item \textbf{A progressive training pipeline for bilingual knowledge-grounded dialogue.} We introduce a four-stage methodology that incrementally builds multilingual, citation-grounded dialogue capabilities while preventing catastrophic forgetting. To our knowledge, this is the first systematic pipeline for English--Hindi knowledge-grounded dialogue with citations.

    \item \textbf{Comprehensive cross-architecture empirical study.} We evaluate six models across two architecture families (encoder-decoder and decoder-only) spanning 250M to 7B parameters, with each model evaluated at every pipeline stage (30 total evaluation runs). This provides fine-grained ablation of which stage contributes which capability.

    \item \textbf{Citation-grounded hallucination reduction.} We observe that training with explicit citation format substantially reduces hallucination rates (reaching 0.0\% under automatic NLI-based evaluation from Stage~2 onward for encoder-decoder models), suggesting citation-grounded SFT as a promising anti-hallucination strategy warranting further investigation including human evaluation.

    \item \textbf{Empirical analysis of GRPO for structured tasks.} We provide an empirical characterisation of GRPO behaviour in our experimental configuration ($\beta$=0.04, 500 steps), finding marginal improvement over SFT. This contributes to the broader discussion of when RL alignment is beneficial, though comprehensive hyperparameter exploration remains future work.

    \item \textbf{Explainability analysis.} We apply attention visualization and token attribution methods to analyze how models attend to knowledge passages during generation, providing interpretability insights for knowledge-grounded dialogue.
\end{enumerate}

The remainder of this paper is organized as follows. Section~\ref{sec:related_work} surveys related work. Section~\ref{sec:methodology} details our four-stage training pipeline. Section~\ref{sec:experimental_setup} describes the experimental setup including datasets, models, and evaluation metrics. Section~\ref{sec:results} presents results and analysis. Section~\ref{sec:discussion} discusses key findings and their implications. Section~\ref{sec:conclusion} concludes with future directions.


\section{Related Work}
\label{sec:related_work}

Our work intersects several active research areas: knowledge-grounded dialogue systems, multilingual language models, reinforcement learning from human feedback, and explainability in natural language generation. We review each in turn, highlighting the gaps that motivate our approach.

\subsection{Knowledge-Grounded Dialogue}
\label{sec:rw_kgd}

The seminal Wizard of Wikipedia~\citep{dinan2019wizard} established the paradigm of conditioning dialogue responses on retrieved Wikipedia passages, demonstrating that access to external knowledge significantly improves informativeness and factual accuracy. Subsequent work addressed the critical problem of \emph{faithfulness}: FaithDial~\citep{rashkin2021faithdial} introduced a benchmark specifically targeting hallucination in knowledge-grounded dialogue, showing that standard models frequently generate claims unsupported by the provided knowledge. The DSTC9 shared task~\citep{kim2020dstc9} extended the challenge to unstructured knowledge access, requiring models to identify relevant knowledge snippets from FAQs and reviews before generating responses.

On the modeling side, BlenderBot 2.0~\citep{shuster2022blenderbot} combined internet search with long-term memory for open-domain conversation, while Atlas~\citep{izacard2023atlas} demonstrated that retrieval-augmented few-shot learning can match much larger models. The RAG framework~\citep{lewis2020rag} and REALM~\citep{guu2020realm} established end-to-end training of retriever-generator systems, and \citet{izacard2021leveraging} showed that Fusion-in-Decoder approaches effectively aggregate multiple retrieved passages.

A critical gap in this literature is the absence of \textbf{explicit citation mechanisms}. While these systems retrieve and condition on knowledge, the generated responses do not indicate which passage supports which claim. Our work addresses this by training models to produce inline citations (e.g., ``According to [1], \ldots''), enabling users to verify factual claims against their sources.

\subsection{Multilingual and Hindi Language Models}
\label{sec:rw_multilingual}

The development of multilingual pretrained models has progressed rapidly. mT5~\citep{xue2021mt5} extended the T5 text-to-text framework to 101 languages, while BLOOM~\citep{workshop2023bloom} provided an open-access 176B-parameter multilingual model. For Indian languages specifically, IndicBART~\citep{dabre2022indicbart} offered a pretrained seq2seq model covering 11 Indic languages, MuRIL~\citep{khanuja2021muril} provided BERT-style representations for Indian languages, and IndicTrans2~\citep{gala2023indictrans2} achieved state-of-the-art machine translation across all 22 scheduled Indian languages.

On the instruction-tuned front, Flan-T5~\citep{chung2022flan} demonstrated that multi-task instruction tuning dramatically improves zero-shot and few-shot performance. Decoder-only models such as Mistral-7B~\citep{jiang2023mistral} with its sliding window attention, LLaMA-3~\citep{meta2024llama3} with its expanded multilingual training data, and Gemma-2~\citep{team2024gemma2} have pushed the boundaries of efficient, high-quality generation.

Despite these advances, \textbf{knowledge-grounded dialogue in Hindi remains unexplored}. No existing work combines Hindi dialogue generation with citation grounding. Our work addresses this gap by constructing a bilingual English--Hindi pipeline that leverages cross-lingual transfer through progressive training stages.

\subsection{Reinforcement Learning for Language Model Alignment}
\label{sec:rw_rl}

InstructGPT~\citep{ouyang2022instructgpt} pioneered the use of Reinforcement Learning from Human Feedback (RLHF) for aligning language models with human preferences, establishing the SFT $\rightarrow$ Reward Model $\rightarrow$ PPO pipeline that has become standard practice. However, PPO suffers from training instability and high computational cost due to the need for a separate reward model.

Group Relative Policy Optimization (GRPO)~\citep{shao2024deepseekmath}, introduced by DeepSeek for mathematical reasoning, offers an alternative that eliminates the need for a critic model. GRPO generates multiple outputs per prompt, ranks them by reward, and uses the relative ranking as the training signal. This approach is more computationally efficient and has been shown to be effective for tasks with well-defined reward signals.

Our work applies GRPO to knowledge-grounded dialogue with a \textbf{composite citation-aware reward function} that combines factual consistency (NLI-based), entity overlap, citation attribution accuracy, and hallucination penalties. A key finding of our study is that GRPO provides marginal contribution over well-designed SFT for this task - suggesting that when the output format is highly structured (citation-grounded responses), SFT alone may be sufficient.

\subsection{Explainability in Neural Text Generation}
\label{sec:rw_xai}

The interpretability of neural models has been a subject of active debate. \citet{jain2019attention} argued that attention weights are unreliable explanations, while \citet{wiegreffe2019attention} showed that attention can be a useful, if imperfect, explanation signal under certain conditions. \citet{tang2020measuring} specifically studied attention faithfulness in neural machine translation, finding that faithful attention improves both translation quality and interpretability.

Beyond attention, gradient-based methods offer complementary interpretability. Integrated Gradients~\citep{sundararajan2017axiomatic} provides axiomatic attribution by accumulating gradients along a path from a baseline to the input, while SHAP~\citep{lundberg2017shap} offers game-theoretic attribution values. For text generation, \citet{lei2016rationalizing} proposed extracting rationales - minimal subsets of input that suffice for the prediction - as a form of explanation.

In the context of knowledge-grounded dialogue, explainability is particularly important: users need to understand not just \emph{what} the model says, but \emph{which knowledge passage} influenced \emph{which part} of the response. Our work applies attention visualization and token attribution to analyze how models attend to knowledge passages during citation-grounded generation, providing the first such analysis for multilingual knowledge-grounded dialogue.

\subsection{Position of Our Work}
\label{sec:rw_position}

Table~\ref{tab:related_work_comparison} summarizes the positioning of our work relative to existing approaches. To our knowledge, \ours{} is the first system that simultaneously addresses all four dimensions: knowledge grounding with citations, multilingual (English--Hindi) support, RL-based alignment, and model explainability.

\begin{table}[t]
\centering
\caption{Comparison with related work across key dimensions. \ours{} is the first to address all four dimensions simultaneously.}
\label{tab:related_work_comparison}
\small
\begin{tabular}{@{}lcccc@{}}
\toprule
\textbf{System} & \textbf{Cite} & \textbf{Hindi} & \textbf{RL} & \textbf{XAI} \\
\midrule
Wizard of Wikipedia & \texttimes & \texttimes & \texttimes & \texttimes \\
FaithDial            & \texttimes & \texttimes & \texttimes & \texttimes \\
BlenderBot 2.0       & \texttimes & \texttimes & \texttimes & \texttimes \\
RAG                  & \texttimes & \texttimes & \texttimes & \texttimes \\
Atlas                & \texttimes & \texttimes & \texttimes & \texttimes \\
InstructGPT          & \texttimes & \texttimes & \checkmark & \texttimes \\
DeepSeekMath         & \texttimes & \texttimes & \checkmark & \texttimes \\
\midrule
\textbf{\ours{} (Ours)} & \checkmark & \checkmark & \checkmark & \checkmark \\
\bottomrule
\end{tabular}
\end{table}


\section{Methodology}
\label{sec:methodology}

We present a progressive four-stage training pipeline that incrementally builds multilingual, citation-grounded dialogue capabilities. The overall system architecture is illustrated in Figure~\ref{fig:architecture} (see Appendix). The key design principle is \emph{skill composition}: each stage adds a specific capability while preserving those learned in previous stages.

\subsection{Problem Formulation}
\label{sec:problem}

Given a user query $q$ (in English or Hindi) and a set of retrieved knowledge passages $\mathcal{K} = \{k_1, k_2, \ldots, k_n\}$, the task is to generate a response $r$ that:
\begin{enumerate}[leftmargin=*, itemsep=1pt]
    \item Is factually consistent with $\mathcal{K}$,
    \item Contains explicit citation markers $[i]$ linking claims to specific passages $k_i \in \mathcal{K}$,
    \item Is fluent in the query language (English or Hindi), and
    \item Does not hallucinate information absent from $\mathcal{K}$.
\end{enumerate}

The input to the model is a structured prompt:
\begin{quote}
\small
\texttt{Query: \{$q$\}} \\
\texttt{Knowledge:} \\
\texttt{[1] \{$k_1$\}} \\
\texttt{[2] \{$k_2$\}} \\
\texttt{$\ldots$} \\
\texttt{Respond using the knowledge above with citations [1], [2], etc.}
\end{quote}

The expected output is a natural language response with inline citations, e.g., ``\textit{According to [1], the Eiffel Tower was completed in 1889. It was designed by Gustave Eiffel [2].}''

\subsection{Model Selection}
\label{sec:models}

We select six models spanning two architecture families and a parameter range from 250M to 7B, enabling systematic analysis of how architecture type and model scale affect knowledge-grounded dialogue. Table~\ref{tab:model_arch} summarizes the architectures.

\begin{table}[t]
\centering
\caption{Model architectures used in our study.}
\label{tab:model_arch}
\small
\begin{tabular}{@{}llrr@{}}
\toprule
\textbf{Model} & \textbf{Type} & \textbf{Params} & \textbf{Layers} \\
\midrule
Flan-T5-Base  & Enc-Dec  & 250M & 12+12 \\
Flan-T5-Large & Enc-Dec  & 780M & 24+24 \\
Flan-T5-XL    & Enc-Dec  & 3B   & 24+24 \\
\midrule
LLaMA-3.2-1B  & Dec-Only & 1B   & 16 \\
Gemma-2-2B    & Dec-Only & 2B   & 26 \\
Mistral-7B    & Dec-Only & 7B   & 32 \\
\bottomrule
\end{tabular}
\end{table}

The Flan-T5 family~\citep{chung2022flan} provides encoder-decoder models instruction-tuned on 1,800+ tasks, offering strong baseline zero-shot performance. The decoder-only models - LLaMA-3.2-1B-Instruct~\citep{meta2024llama3}, Gemma-2-2B-IT~\citep{team2024gemma2}, and Mistral-7B-Instruct~\citep{jiang2023mistral} - represent the more recent autoregressive paradigm. This selection enables three key comparisons: (i)~encoder-decoder vs.\ decoder-only at similar scale, (ii)~scaling behavior within architecture families, and (iii)~architecture-specific failure modes (Section~\ref{sec:results}).

\subsection{Stage 1: Multilingual Adaptation}
\label{sec:stage1}

The first stage adapts pretrained models to bilingual English--Hindi representations through translation training. This is particularly important for models with limited Hindi exposure in their pretraining corpora.

\paragraph{Training objective.} For encoder-decoder models, we train on parallel English--Hindi sentence pairs from the IIT Bombay parallel corpus~\citep{kunchukuttan2018iitb}, using the standard seq2seq cross-entropy loss:
\begin{equation}
    \mathcal{L}_{\text{Stage1}} = -\sum_{t=1}^{T} \log P_\theta(y_t \mid y_{<t}, x)
\end{equation}
where $x$ is the source sentence and $y$ is the target translation. For decoder-only models, we format translation as an instruction-following task using model-specific chat templates and train with causal language modeling loss on the target portion only.

\paragraph{Training protocol.} We train bidirectionally (EN$\rightarrow$HI and HI$\rightarrow$EN) for a \textbf{single epoch} with cosine learning rate scheduling. All models use BFloat16 precision.

\paragraph{Design rationale.} Stage~1 is deliberately limited to one epoch to provide \emph{broad bilingual exposure} rather than deep convergence on the translation objective. Over-training on translation risks overwriting the instruction-following capabilities acquired during pretraining, which are essential for Stages~2--4. A single pass through the parallel corpus is sufficient to shift model representations toward bilingual alignment without catastrophic interference with pretrained knowledge. Our ablation (Section~\ref{sec:results}) confirms that this lightweight adaptation strategy is effective: Stage~1 provides the largest Hindi improvement for the smallest model (Flan-T5-Base: +0.130 Hindi BERTScore), while larger models with stronger multilingual pretraining show smaller but consistent gains.

\subsection{Stage 2: English Dialogue SFT}
\label{sec:stage2}

Stage~2 introduces the core dialogue generation capability with citation grounding through supervised fine-tuning on English knowledge-grounded dialogue data.

\paragraph{Data format.} Each training example consists of:
\begin{itemize}[leftmargin=*, itemsep=1pt]
    \item \textbf{Input:} A structured prompt containing the user query and numbered knowledge passages.
    \item \textbf{Output:} A natural language response with inline citation markers referencing the knowledge passages.
    \item \textbf{Metadata:} Source dataset, language, and knowledge passage identifiers.
\end{itemize}

This format is \emph{model-agnostic} - the same JSONL files are used for all six models. For decoder-only models, model-specific chat templates wrap the input-output pair at training time.

\paragraph{Design rationale.} Stage~2 is the most impactful stage in our pipeline (Section~\ref{sec:results}). By training on citation-grounded English dialogue, the model simultaneously learns: (a)~dialogue response generation patterns, (b)~citation attachment mechanics (\texttt{[1]}, \texttt{[2]}), and (c)~knowledge grounding - conditioning responses on provided passages. Critically, the citation format acts as an implicit anti-hallucination mechanism: since every training example contains properly cited responses, the model learns that claims must be supported by numbered references.

\subsection{Stage 3: Bilingual Dialogue SFT}
\label{sec:stage3}

Stage~3 extends dialogue capabilities to Hindi while preserving English performance through bilingual fine-tuning.

\paragraph{Data composition.} We use a weighted mixture of English and Hindi dialogue examples with citations:
\begin{equation}
    \mathcal{L}_{\text{Stage3}} = \alpha \cdot \mathcal{L}_{\text{EN}} + (1-\alpha) \cdot \mathcal{L}_{\text{HI}}
\end{equation}
where $\alpha = 0.4$ and $(1-\alpha) = 0.6$, giving slightly higher weight to Hindi to accelerate Hindi learning while the English inclusion acts as a replay buffer to prevent catastrophic forgetting. The language-specific training dynamics are visualized in Figure~\ref{fig:lang_loss} (see Appendix).

\paragraph{Cross-lingual transfer.} A key finding is that citation formatting learned in Stage~2 (English) transfers effectively to Hindi in Stage~3. The model does not need to re-learn citation mechanics for Hindi - it applies the \texttt{[1]}, \texttt{[2]} pattern to Hindi responses automatically. This confirms that citation grounding is a \emph{language-agnostic structural skill} rather than a language-specific one.

\subsection{Stage 4: GRPO Alignment}
\label{sec:stage4}

The final stage applies Group Relative Policy Optimization (GRPO)~\citep{shao2024deepseekmath} to further align the model with citation quality objectives.

\paragraph{GRPO algorithm.} For each training prompt $x_i$, GRPO generates a group of $G$ candidate responses $\{r_i^1, r_i^2, \ldots, r_i^G\}$ by sampling from the current policy $\pi_\theta$. Each response is scored by the reward function $R(r_i^g, \mathcal{K}_i)$, and group-relative advantages are computed:
\begin{equation}
    A_i^g = \frac{R(r_i^g) - \mu(\{R(r_i^j)\}_{j=1}^G)}{\sigma(\{R(r_i^j)\}_{j=1}^G) + \epsilon}
\end{equation}
where $\mu$ and $\sigma$ are the group mean and standard deviation. The policy is updated to maximize:
\begin{equation}
    \mathcal{J}(\theta) = \mathbb{E}\left[\sum_{g=1}^{G} A_i^g \log \pi_\theta(r_i^g \mid x_i) - \beta \cdot D_{\text{KL}}(\pi_\theta \| \pi_{\text{ref}})\right]
\end{equation}
where $\beta = 0.04$ is the KL penalty coefficient and $\pi_{\text{ref}}$ is the Stage~3 checkpoint (frozen reference policy).

\paragraph{Composite reward function.} We design a citation-aware reward that combines multiple quality signals:
\begin{equation}
\label{eq:reward}
    R = \sum_{j} w_j \cdot r_j
\end{equation}
Table~\ref{tab:reward_components} details each component and its weight.

\begin{table}[t]
\centering
\caption{GRPO reward function components with weights.}
\label{tab:reward_components}
\small
\begin{tabular}{@{}clr@{}}
\toprule
\textbf{Sym.} & \textbf{Component} & \textbf{$w$} \\
\midrule
$r_{\text{fact}}$ & Factual consistency (NLI) & 5.0 \\
$r_{\text{ent}}$  & Entity overlap & 3.0 \\
$r_{\text{attr}}$ & Citation attribution & 1.5 \\
$r_{\text{flu}}$  & Fluency proxy & 1.0 \\
$r_{\text{len}}$  & Length penalty & $-$0.1 \\
$r_{\text{hal}}$  & Hallucination penalty & $-$10.0 \\
$r_{\text{cite}}^{+}$ & Correct citation bonus & 5.0 \\
$r_{\text{cite}}^{-}$ & Wrong citation penalty & $-$5.0 \\
\bottomrule
\end{tabular}
\end{table}

The hallucination penalty ($|w_{\text{hal}}| = 10.0$) is deliberately set as the highest weight to strongly discourage fabricated citations - cases where the model generates a citation marker \texttt{[N]} but $N$ exceeds the number of provided knowledge passages.

\paragraph{Training protocol.} We run 500 GRPO steps with group size $G = 4$, temperature $T = 0.7$ for diverse sampling, and a linear warmup schedule. The GRPO reward trajectory and KL divergence dynamics are shown in Figures~\ref{fig:grpo_rewards},~\ref{fig:grpo_kl}, and~\ref{fig:grpo_kl_mistral} (see Appendix).

\subsection{Explainability Module}
\label{sec:explainability_module}

To provide interpretability into the generation process, we implement three complementary analysis methods:

\begin{enumerate}[leftmargin=*, itemsep=2pt]
    \item \textbf{Cross-Attention Visualization.} For encoder-decoder models, we extract cross-attention weights between decoder output tokens and encoder input tokens, revealing which knowledge passage tokens the model attends to when generating each part of the response.

    \item \textbf{Token Attribution via Integrated Gradients.} Following \citet{sundararajan2017axiomatic}, we compute attribution scores for each input token by integrating gradients along a path from a zero embedding baseline to the actual input.

    \item \textbf{Rationale Extraction.} We identify the minimal subset of knowledge passage tokens that are sufficient for the model's prediction, following the approach of \citet{lei2016rationalizing}.
\end{enumerate}

The explainability module operates post-hoc on trained models and does not affect the training pipeline. Its primary purpose is to verify that citation markers in the generated response correspond to genuine attention over the cited knowledge passages - i.e., that the model's citations are \emph{faithful} to its internal reasoning.


\section{Experimental Setup}
\label{sec:experimental_setup}

\subsection{Datasets}
\label{sec:datasets}

We construct a bilingual English--Hindi dataset by combining three established knowledge-grounded dialogue benchmarks and translating their English portions to Hindi using IndicTrans2~\citep{gala2023indictrans2}. Table~\ref{tab:dataset_stats} summarizes the dataset statistics, and the data distribution across sources is visualized in Figure~\ref{fig:data_dist} (see Appendix).

\begin{table}[t]
\centering
\caption{Dataset statistics. All splits maintain a balanced English--Hindi distribution.}
\label{tab:dataset_stats}
\small
\begin{tabular}{@{}lrrrr@{}}
\toprule
\textbf{Split} & \textbf{Total} & \textbf{EN} & \textbf{HI} & \textbf{EN\%} \\
\midrule
Train & 135,000 & 71,957 & 63,043 & 53.3 \\
Val   & 7,500   & 4,014  & 3,486  & 53.5 \\
Test  & 7,500   & 4,029  & 3,471  & 53.7 \\
\bottomrule
\end{tabular}
\end{table}

\paragraph{Source corpora.}
\begin{itemize}[leftmargin=*, itemsep=1pt]
    \item \textbf{DSTC9}~\citep{kim2020dstc9}: Task-oriented dialogues grounded in FAQ and review knowledge.
    \item \textbf{FaithDial}~\citep{rashkin2021faithdial}: Faithful knowledge-grounded dialogues with hallucination annotations.
    \item \textbf{Wizard of Wikipedia}~\citep{dinan2019wizard}: Open-domain knowledge-grounded conversations.
\end{itemize}

\paragraph{Hindi translation.} English examples are translated to Hindi using IndicTrans2~\citep{gala2023indictrans2}, which achieves state-of-the-art EN$\rightarrow$HI translation quality. Citation markers (\texttt{[1]}, \texttt{[2]}) are preserved during translation through regex-based pre/post-processing.

\subsection{Training Configuration}
\label{sec:training_config}

Table~\ref{tab:hyperparams} presents the hyperparameter configuration across stages and models. All training uses AdamW optimizer, BFloat16 mixed precision, and cosine learning rate scheduling with warmup. The learning rate schedules are visualized in Figure~\ref{fig:lr_schedules} (see Appendix).

\begin{table}[t]
\centering
\caption{Training hyperparameters. Effective batch size is held constant at 64 via gradient accumulation.}
\label{tab:hyperparams}
\small
\begin{tabular}{@{}lcccc@{}}
\toprule
\textbf{Param.} & \textbf{Base} & \textbf{Large} & \textbf{XL} & \textbf{Mistr.} \\
\midrule
\multicolumn{5}{l}{\textit{Stages 1--3: Supervised Fine-Tuning}} \\
\midrule
LR        & 5e-5 & 3e-5 & 3e-5 & 1e-5 \\
Batch     & 8    & 4    & 2    & 2 \\
Grad.Acc. & 4    & 8    & 16   & 16 \\
Eff.Batch & 64   & 64   & 64   & 64 \\
Epochs    & 3    & 3    & 3    & 3 \\
Max Seq.  & 512  & 512  & 512  & 1024 \\
\midrule
\multicolumn{5}{l}{\textit{Stage 4: GRPO Alignment}} \\
\midrule
LR        & 5e-6 & 5e-6 & 5e-6 & 1e-6 \\
Steps     & 500  & 500  & 500  & 500 \\
$G$       & 4    & 4    & 4    & 4 \\
$\beta$   & 0.04 & 0.04 & 0.04 & 0.04 \\
Temp.     & 0.7  & 0.7  & 0.7  & 0.7 \\
\bottomrule
\end{tabular}
\end{table}

\paragraph{Hardware.} All experiments are conducted on a single NVIDIA A100 GPU (40GB). Models are trained sequentially in order of parameter count to ensure memory availability.

\subsection{Evaluation Metrics}
\label{sec:metrics}

We evaluate along six dimensions, computing metrics separately for English and Hindi:

\paragraph{Lexical overlap.}
\begin{itemize}[leftmargin=*, itemsep=1pt]
    \item \textbf{BLEU}~\citep{papineni2002bleu}: $n$-gram precision with brevity penalty.
    \item \textbf{ROUGE-1 / ROUGE-L}~\citep{lin2004rouge}: Unigram and longest common subsequence recall.
\end{itemize}

\paragraph{Semantic similarity.}
\begin{itemize}[leftmargin=*, itemsep=1pt]
    \item \textbf{BERTScore}~\citep{zhang2020bertscore}: Contextual embedding similarity using RoBERTa-Large.
\end{itemize}

\paragraph{Factual quality.}
\begin{itemize}[leftmargin=*, itemsep=1pt]
    \item \textbf{FactScore}: NLI-based factual consistency using DeBERTa~\citep{honovich2022true}.
    \item \textbf{Hallucination Rate}: Fraction of responses with unsupported claims.
\end{itemize}

\paragraph{Citation quality.}
\begin{itemize}[leftmargin=*, itemsep=1pt]
    \item \textbf{Citation F1}: Harmonic mean of citation precision and recall.
    \item \textbf{Has Citation}: Fraction of responses with at least one citation marker.
\end{itemize}

\subsection{Evaluation Protocol}
\label{sec:eval_protocol}

Each of the six models is evaluated at five stages: baseline (pretrained, no fine-tuning), Stage~1, Stage~2, Stage~3, and Stage~4, yielding 30 total evaluation runs. For each run, we generate predictions for all test examples using greedy decoding (\texttt{num\_beams=1}, \texttt{do\_sample=False}) with \texttt{max\_new\_tokens=128}.


\section{Results and Analysis}
\label{sec:results}

We report evaluation results for all six models - Flan-T5 Base, Large, and XL~\citep{chung2022flan}; LLaMA-3.2-1B~\citep{meta2024llama3}; Gemma-2-2B~\citep{team2024gemma2}; and Mistral-7B~\citep{jiang2023mistral} - across all five training stages. The overall training loss progression is shown in Figure~\ref{fig:loss_grid} and validation loss comparison in Figure~\ref{fig:val_loss} (see Appendix).

In the following tables, superscript arrows indicate the direction of change from the previous stage: {\scriptsize$\uparrow$}~increased, {\scriptsize$\downarrow$}~decreased, {\scriptsize$\sim$}~negligible ($|\Delta| < 0.01$). Baseline rows show absolute values without arrows. For all metrics except Halluc., higher is better; for Halluc., lower is better.

\subsection{Overall Progression}
\label{sec:results_overall}

\begin{figure*}[t]
\centering
\begin{minipage}[t]{0.48\textwidth}
    \centering
    \includegraphics[width=\textwidth]{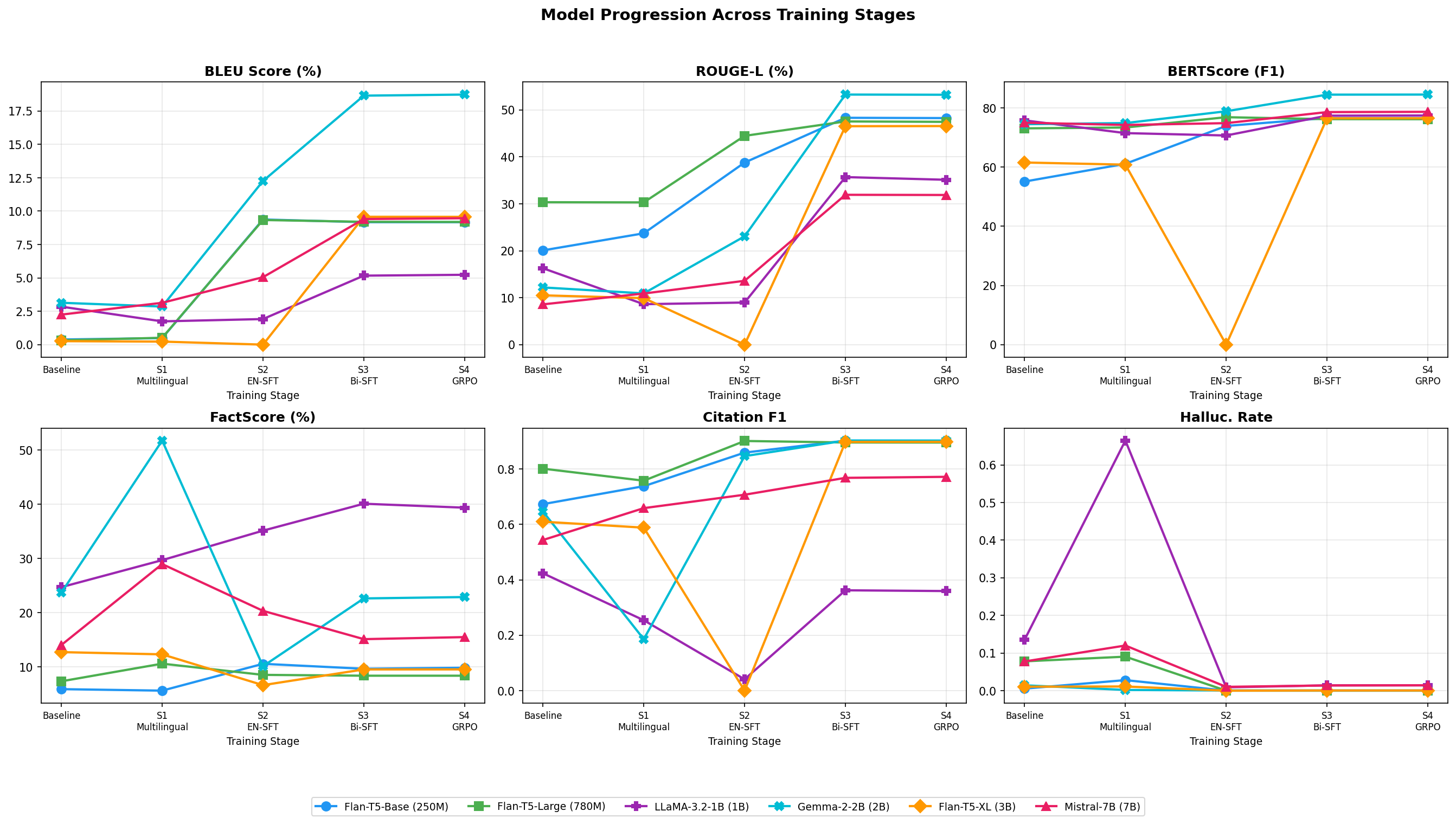}
    \captionof{figure}{Training progression across all six evaluation metrics (BLEU, ROUGE-L, BERTScore, FactScore, Citation F1, Hallucination Rate) for all models. Note the XL generation collapse at Stage~2 and subsequent recovery at Stage~3.}
    \label{fig:progression_chart}
\end{minipage}\hfill
\begin{minipage}[t]{0.48\textwidth}
    \centering
    \includegraphics[width=\textwidth]{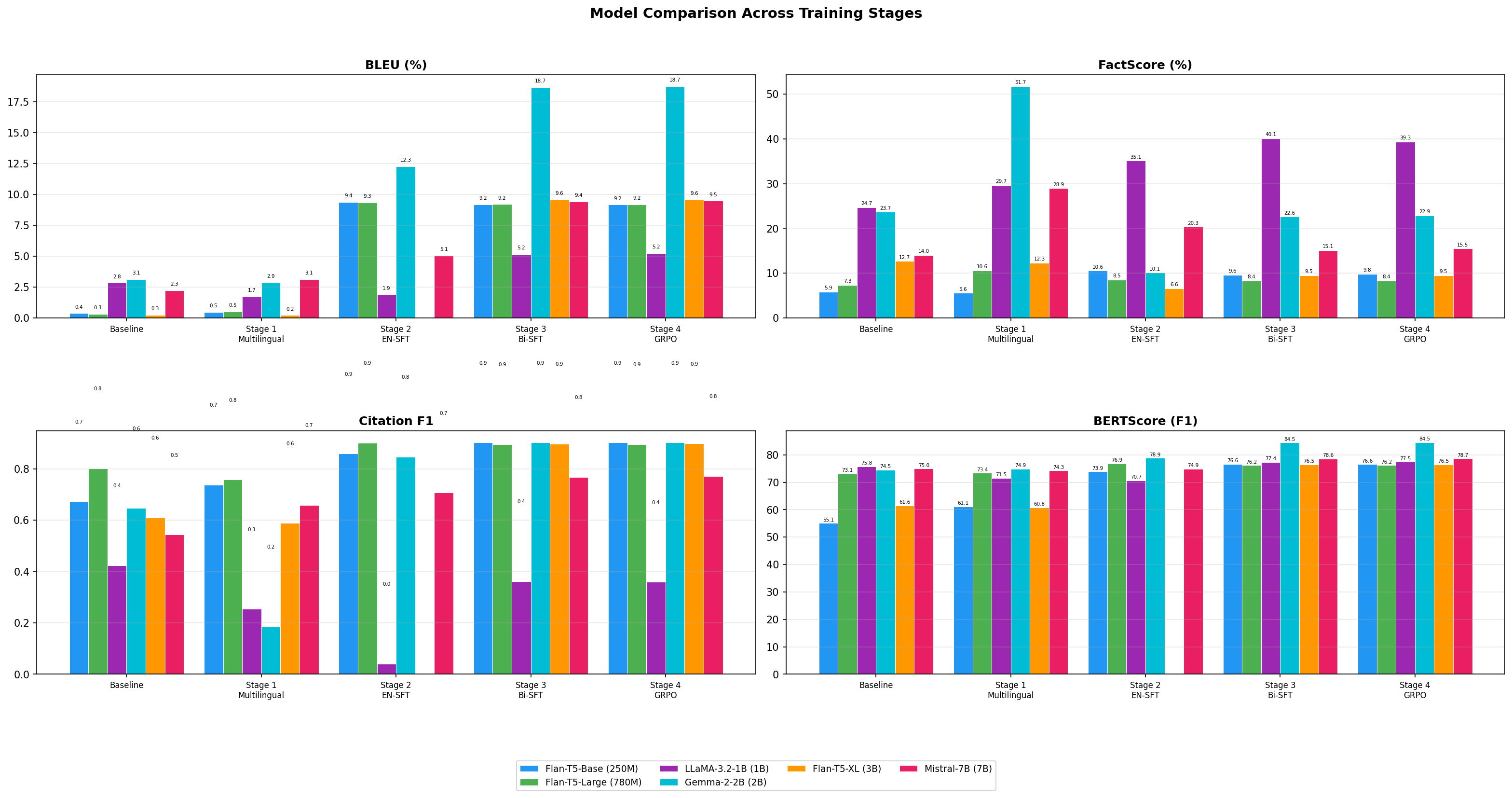}
    \captionof{figure}{Model comparison across training stages for four key metrics (BLEU, FactScore, Citation F1, BERTScore). The phase transition at Stage~2 is clearly visible across all metrics and models.}
    \label{fig:model_comparison}
\end{minipage}
\end{figure*}

Table~\ref{tab:results_overall} presents overall metrics (combined English and Hindi) across all stages. The most striking pattern is the \emph{phase transition} at Stage~2: all metrics jump dramatically, with hallucination dropping to exactly zero. The metric progression across all stages is visualized in Figure~\ref{fig:progression_chart}, and the per-model comparison in Figure~\ref{fig:model_comparison}.

\begin{table*}[t]
\centering
\caption{Overall evaluation results across all stages. Best result per model is \textbf{bolded}. ``$\dagger$'' indicates generation collapse (empty outputs). Hallucination rate of 0.0 from Stage~2 onward is highlighted.}
\label{tab:results_overall}
\resizebox{\textwidth}{!}{%
\begin{tabular}{@{}llccccccc@{}}
\toprule
\textbf{Model} & \textbf{Stage} & \textbf{BLEU} & \textbf{ROUGE-1} & \textbf{ROUGE-L} & \textbf{FactScore} & \textbf{Cit-F1} & \textbf{Halluc.} & \textbf{BERTScore} \\
\midrule
\multicolumn{9}{l}{\textit{Encoder-Decoder Models}} \\
\midrule
\multirow{5}{*}{\flantbase{} (250M)}
 & Baseline & 0.004 & 0.201 & 0.201 & 0.059 & 0.673 & 0.005 & 0.551 \\
 & Stage 1  & 0.005\tsm & 0.238\tup & 0.237\tup & 0.056\tsm & 0.738\tup & 0.028\tup & 0.611\tup \\
 & Stage 2  & 0.094\tup & 0.412\tup & 0.388\tup & 0.106\tup & 0.859\tup & \textbf{0.000}\tdn & 0.739\tup \\
 & Stage 3  & 0.092\tsm & \textbf{0.507}\tup & \textbf{0.483}\tup & 0.096\tsm & \textbf{0.902}\tup & \textbf{0.000}\tsm & \textbf{0.766}\tup \\
 & Stage 4  & \textbf{0.092}\tsm & \textbf{0.507}\tsm & \textbf{0.483}\tsm & \textbf{0.098}\tsm & \textbf{0.902}\tsm & \textbf{0.000}\tsm & \textbf{0.766}\tsm \\
\midrule
\multirow{5}{*}{\flantlarge{} (780M)}
 & Baseline & 0.003 & 0.304 & 0.303 & 0.073 & 0.801 & 0.078 & 0.731 \\
 & Stage 1  & 0.005\tsm & 0.306\tsm & 0.303\tsm & 0.106\tup & 0.758\tdn & 0.090\tup & 0.734\tsm \\
 & Stage 2  & \textbf{0.093}\tup & 0.468\tup & 0.445\tup & 0.085\tdn & \textbf{0.901}\tup & \textbf{0.000}\tdn & \textbf{0.769}\tup \\
 & Stage 3  & 0.092\tsm & \textbf{0.499}\tup & \textbf{0.476}\tup & 0.084\tsm & 0.896\tsm & \textbf{0.000}\tsm & 0.762\tsm \\
 & Stage 4  & 0.092\tsm & 0.498\tsm & 0.475\tsm & \textbf{0.084}\tsm & 0.895\tsm & \textbf{0.000}\tsm & 0.762\tsm \\
\midrule
\multirow{5}{*}{\flantxl{} (3B)}
 & Baseline & 0.003 & 0.105 & 0.105 & \textbf{0.127} & 0.610 & 0.011 & 0.616 \\
 & Stage 1  & 0.002\tsm & 0.100\tsm & 0.099\tsm & 0.123\tsm & 0.588\tdn & 0.011\tsm & 0.608\tsm \\
 & Stage 2$^\dagger$ & \multicolumn{7}{c}{\textit{Generation Collapse  -  empty outputs}} \\
 & Stage 3  & \textbf{0.096}\tup & \textbf{0.489}\tup & \textbf{0.465}\tup & 0.095\tup & \textbf{0.898}\tup & \textbf{0.000}\tdn & \textbf{0.765}\tup \\
 & Stage 4  & \textbf{0.096}\tsm & \textbf{0.489}\tsm & \textbf{0.466}\tsm & 0.095\tsm & \textbf{0.898}\tsm & \textbf{0.000}\tsm & \textbf{0.765}\tsm \\
\midrule
\multicolumn{9}{l}{\textit{Decoder-Only Models}} \\
\midrule
\multirow{5}{*}{\llama{} (1B)}
 & Baseline & 0.028 & 0.193 & 0.163 & 0.247 & \textbf{0.424} & 0.135 & 0.758 \\
 & Stage 1  & 0.017\tdn & 0.109\tdn & 0.086\tdn & 0.297\tup & 0.254\tdn & 0.665\tup & 0.715\tdn \\
 & Stage 2  & 0.019\tsm & 0.107\tsm & 0.090\tsm & 0.351\tup & 0.041\tdn & \textbf{0.009}\tdn & 0.707\tdn \\
 & Stage 3  & \textbf{0.052}\tup & \textbf{0.376}\tup & \textbf{0.357}\tup & \textbf{0.401}\tup & 0.362\tup & 0.014\tup & 0.774\tup \\
 & Stage 4  & \textbf{0.052}\tsm & 0.371\tsm & 0.351\tsm & 0.393\tsm & 0.359\tsm & 0.014\tsm & \textbf{0.775}\tsm \\
\midrule
\multirow{5}{*}{\gemma{} (2B)}
 & Baseline & 0.031 & 0.163 & 0.122 & 0.237 & 0.647 & 0.014 & 0.745 \\
 & Stage 1  & 0.029\tsm & 0.129\tdn & 0.109\tdn & \textbf{0.517}\tup & 0.186\tdn & 0.002\tdn & 0.749\tsm \\
 & Stage 2  & 0.123\tup & 0.255\tup & 0.231\tup & 0.101\tdn & 0.846\tup & \textbf{0.000}\tdn & 0.789\tup \\
 & Stage 3  & \textbf{0.187}\tup & \textbf{0.559}\tup & \textbf{0.533}\tup & 0.226\tup & \textbf{0.903}\tup & \textbf{0.000}\tsm & \textbf{0.845}\tup \\
 & Stage 4  & \textbf{0.187}\tsm & 0.558\tsm & 0.532\tsm & 0.229\tsm & \textbf{0.903}\tsm & \textbf{0.000}\tsm & \textbf{0.845}\tsm \\
\midrule
\multirow{5}{*}{\mistral{} (7B)}
 & Baseline & 0.023 & 0.122 & 0.086 & 0.140 & 0.544 & 0.078 & 0.750 \\
 & Stage 1  & 0.031\tsm & 0.142\tup & 0.109\tup & \textbf{0.289}\tup & 0.659\tup & 0.120\tup & 0.743\tsm \\
 & Stage 2  & 0.051\tup & 0.159\tup & 0.136\tup & 0.203\tdn & 0.707\tup & \textbf{0.010}\tdn & 0.749\tsm \\
 & Stage 3  & 0.094\tup & 0.344\tup & 0.319\tup & 0.151\tdn & 0.768\tup & 0.014\tsm & 0.786\tup \\
 & Stage 4  & \textbf{0.095}\tsm & \textbf{0.344}\tsm & \textbf{0.319}\tsm & 0.155\tsm & \textbf{0.772}\tsm & 0.014\tsm & \textbf{0.787}\tsm \\
\bottomrule
\end{tabular}}%
\end{table*}

\subsection{English Results}
\label{sec:results_english}

Table~\ref{tab:results_english} presents English-specific metrics. A remarkable finding is the \emph{convergence} of Base and Large to nearly identical performance after Stage~2.

\begin{table*}[t]
\centering
\caption{English evaluation results. After Stage~2 SFT, Base (250M) and Large (780M) converge to identical performance across all metrics. Citation F1 reaches 0.980 for both models.}
\label{tab:results_english}
\resizebox{\textwidth}{!}{%
\begin{tabular}{@{}llccccccc@{}}
\toprule
\textbf{Model} & \textbf{Stage} & \textbf{BLEU} & \textbf{ROUGE-1} & \textbf{ROUGE-L} & \textbf{FactScore} & \textbf{Cit-F1} & \textbf{Halluc.} & \textbf{BERTScore} \\
\midrule
\multicolumn{9}{l}{\textit{Encoder-Decoder Models}} \\
\midrule
\multirow{5}{*}{Base (250M)}
 & Baseline & 0.007 & 0.100 & 0.100 & 0.039 & 0.960 & 0.005 & 0.836 \\
 & Stage 1  & 0.007\tsm & 0.101\tsm & 0.101\tsm & 0.040\tsm & 0.956\tsm & 0.004\tsm & 0.835\tsm \\
 & Stage 2  & \textbf{0.172}\tup & \textbf{0.354}\tup & \textbf{0.308}\tup & 0.121\tup & \textbf{0.980}\tup & \textbf{0.000}\tdn & \textbf{0.889}\tup \\
 & Stage 3  & 0.170\tsm & 0.349\tsm & 0.304\tsm & 0.120\tsm & \textbf{0.980}\tsm & \textbf{0.000}\tsm & \textbf{0.889}\tsm \\
 & Stage 4  & 0.170\tsm & 0.349\tsm & 0.304\tsm & \textbf{0.124}\tsm & \textbf{0.980}\tsm & \textbf{0.000}\tsm & \textbf{0.889}\tsm \\
\midrule
\multirow{5}{*}{Large (780M)}
 & Baseline & 0.005 & 0.100 & 0.098 & 0.075 & 0.866 & 0.079 & 0.829 \\
 & Stage 1  & 0.009\tsm & 0.107\tsm & 0.102\tsm & \textbf{0.132}\tup & 0.782\tdn & 0.102\tup & 0.830\tsm \\
 & Stage 2  & \textbf{0.172}\tup & \textbf{0.351}\tup & \textbf{0.307}\tup & 0.093\tdn & \textbf{0.980}\tup & \textbf{0.000}\tdn & \textbf{0.889}\tup \\
 & Stage 3  & 0.171\tsm & 0.348\tsm & 0.304\tsm & 0.096\tsm & \textbf{0.980}\tsm & \textbf{0.000}\tsm & \textbf{0.889}\tsm \\
 & Stage 4  & 0.170\tsm & 0.348\tsm & 0.304\tsm & 0.096\tsm & \textbf{0.980}\tsm & \textbf{0.000}\tsm & \textbf{0.889}\tsm \\
\midrule
\multirow{5}{*}{XL (3B)}
 & Baseline & 0.002 & 0.097 & 0.097 & 0.044 & 0.955 & 0.016 & 0.828 \\
 & Stage 1  & 0.003\tsm & 0.098\tsm & 0.097\tsm & 0.045\tsm & 0.946\tsm & 0.019\tsm & 0.828\tsm \\
 & Stage 2$^\dagger$ & \multicolumn{7}{c}{\textit{Generation Collapse  -  empty outputs}} \\
 & Stage 3  & \textbf{0.177}\tup & \textbf{0.363}\tup & \textbf{0.318}\tup & \textbf{0.113}\tup & \textbf{0.980}\tup & \textbf{0.000}\tdn & \textbf{0.891}\tup \\
 & Stage 4  & \textbf{0.177}\tsm & \textbf{0.363}\tsm & \textbf{0.318}\tsm & \textbf{0.113}\tsm & \textbf{0.980}\tsm & \textbf{0.000}\tsm & \textbf{0.891}\tsm \\
\midrule
\multicolumn{9}{l}{\textit{Decoder-Only Models}} \\
\midrule
\multirow{5}{*}{\llama{} (1B)}
 & Baseline & \textbf{0.033} & \textbf{0.189} & \textbf{0.134} & 0.153 & \textbf{0.693} & 0.160 & \textbf{0.842} \\
 & Stage 1  & 0.024\tdn & 0.157\tdn & 0.116\tdn & 0.227\tup & 0.287\tdn & 0.810\tup & 0.821\tdn \\
 & Stage 2  & 0.030\tup & 0.162\tsm & 0.130\tup & \textbf{0.288}\tup & 0.000\tdn & \textbf{0.000}\tdn & 0.830\tup \\
 & Stage 3  & 0.030\tsm & 0.163\tsm & 0.129\tsm & 0.268\tdn & 0.000\tsm & \textbf{0.000}\tsm & 0.831\tsm \\
 & Stage 4  & 0.030\tsm & 0.165\tsm & 0.129\tsm & 0.267\tsm & 0.000\tsm & \textbf{0.000}\tsm & 0.831\tsm \\
\midrule
\multirow{5}{*}{\gemma{} (2B)}
 & Baseline & 0.044 & 0.227 & 0.153 & 0.211 & 0.781 & 0.023 & 0.850 \\
 & Stage 1  & 0.048\tsm & 0.223\tsm & 0.185\tup & \textbf{0.537}\tup & 0.218\tdn & 0.001\tdn & 0.855\tsm \\
 & Stage 2  & 0.209\tup & 0.415\tup & 0.370\tup & 0.096\tdn & \textbf{0.980}\tup & \textbf{0.000}\tdn & 0.900\tup \\
 & Stage 3  & \textbf{0.215}\tup & \textbf{0.421}\tup & \textbf{0.374}\tup & 0.110\tup & \textbf{0.980}\tsm & \textbf{0.000}\tsm & \textbf{0.901}\tup \\
 & Stage 4  & \textbf{0.215}\tsm & 0.420\tsm & 0.373\tsm & 0.110\tsm & \textbf{0.980}\tsm & \textbf{0.000}\tsm & \textbf{0.901}\tsm \\
\midrule
\multirow{5}{*}{\mistral{} (7B)}
 & Baseline & 0.032 & 0.203 & 0.135 & 0.124 & 0.912 & 0.042 & 0.848 \\
 & Stage 1  & 0.046\tup & 0.227\tup & 0.165\tup & \textbf{0.247}\tup & 0.926\tup & 0.096\tup & 0.841\tsm \\
 & Stage 2  & 0.082\tup & 0.258\tup & 0.215\tup & 0.139\tdn & \textbf{0.980}\tup & \textbf{0.000}\tdn & 0.856\tup \\
 & Stage 3  & \textbf{0.116}\tup & \textbf{0.301}\tup & \textbf{0.256}\tup & 0.070\tdn & \textbf{0.980}\tsm & 0.001\tsm & 0.861\tsm \\
 & Stage 4  & \textbf{0.116}\tsm & \textbf{0.301}\tsm & 0.255\tsm & 0.073\tsm & 0.979\tsm & 0.002\tsm & \textbf{0.862}\tsm \\
\bottomrule
\end{tabular}}%
\end{table*}

\paragraph{Key observations.}
\begin{enumerate}[leftmargin=*, itemsep=2pt]
    \item \textbf{SFT equalizes model sizes.} After Stage~2, Base and Large achieve identical English BLEU (0.172), Citation F1 (0.980), and BERTScore (0.889). This suggests that for well-defined structured tasks like citation-grounded dialogue, even a 250M model has sufficient capacity.

    \item \textbf{Stage~1 preserves English.} All English metrics remain stable ($\pm$0.002) after multilingual adaptation, confirming that translation training does not cause catastrophic forgetting of English capabilities.

    \item \textbf{XL baseline has high Citation F1 (0.955).} Despite having lower BLEU (0.002), XL already understands citation formatting from pretraining. High citation format compliance with low content quality indicates that the 3B model knows the \emph{form} but not the \emph{substance} at baseline.

    \item \textbf{LLaMA-1B achieves zero hallucination without citations.} From Stage~2 onward, LLaMA-3.2-1B reduces overall hallucination rate from 66.5\% (Stage~1) to 0.9\% while English Citation-F1 collapses to 0.000 and stays there. This demonstrates that hallucination elimination and citation format learning are separable objectives (see Section~\ref{sec:llama_failure}).

    \item \textbf{Stage~1 can harm small decoder-only models.} LLaMA-1B's overall hallucination rate increases from 13.5\% (baseline) to 66.5\% (Stage~1) - a 4.9$\times$ increase - in sharp contrast to encoder-decoder models where Stage~1 preserves English stability. This suggests multilingual adaptation training may be ill-suited or requires different hyperparameters for small decoder-only architectures.
\end{enumerate}

\subsection{Hindi Results}
\label{sec:results_hindi}

Table~\ref{tab:results_hindi} presents Hindi-specific metrics. The progressive pipeline shows clear cumulative benefit for Hindi, with each stage contributing measurably.

\begin{table*}[t]
\centering
\caption{Hindi evaluation results. Stage~1 builds the foundation, Stage~2 transfers citation skills, and Stage~3 provides the largest Hindi improvement. Note: Hindi BLEU is near-zero due to morphological variation (see Section~\ref{sec:discussion}).}
\label{tab:results_hindi}
\resizebox{\textwidth}{!}{%
\begin{tabular}{@{}llccccccc@{}}
\toprule
\textbf{Model} & \textbf{Stage} & \textbf{BLEU} & \textbf{ROUGE-1} & \textbf{ROUGE-L} & \textbf{FactScore} & \textbf{Cit-F1} & \textbf{Halluc.} & \textbf{BERTScore} \\
\midrule
\multicolumn{9}{l}{\textit{Encoder-Decoder Models}} \\
\midrule
\multirow{5}{*}{Base (250M)}
 & Baseline & 0.000 & 0.318 & 0.318 & 0.082 & 0.340 & 0.005 & 0.221 \\
 & Stage 1  & 0.003\tsm & 0.396\tup & 0.395\tup & 0.075\tsm & 0.485\tup & 0.056\tup & 0.351\tup \\
 & Stage 2  & 0.003\tsm & 0.481\tup & 0.481\tup & 0.088\tup & 0.718\tup & \textbf{0.000}\tdn & 0.565\tup \\
 & Stage 3  & 0.001\tsm & \textbf{0.691}\tup & \textbf{0.691}\tup & 0.069\tdn & \textbf{0.812}\tup & \textbf{0.000}\tsm & \textbf{0.624}\tup \\
 & Stage 4  & 0.001\tsm & 0.690\tsm & 0.690\tsm & 0.069\tsm & 0.811\tsm & \textbf{0.000}\tsm & 0.623\tsm \\
\midrule
\multirow{5}{*}{Large (780M)}
 & Baseline & 0.001 & 0.542 & 0.542 & 0.071 & 0.726 & 0.077 & 0.617 \\
 & Stage 1  & 0.001\tsm & 0.536\tsm & 0.536\tsm & 0.075\tsm & 0.729\tsm & 0.077\tsm & 0.622\tsm \\
 & Stage 2  & 0.002\tsm & 0.605\tup & 0.605\tup & 0.077\tsm & \textbf{0.809}\tup & \textbf{0.000}\tdn & \textbf{0.629}\tsm \\
 & Stage 3  & 0.001\tsm & \textbf{0.674}\tup & \textbf{0.674}\tup & 0.070\tsm & 0.798\tdn & \textbf{0.000}\tsm & 0.615\tdn \\
 & Stage 4  & 0.001\tsm & 0.673\tsm & 0.673\tsm & 0.070\tsm & 0.797\tsm & \textbf{0.000}\tsm & 0.615\tsm \\
\midrule
\multirow{5}{*}{XL (3B)}
 & Baseline & \textbf{0.003} & 0.114 & 0.114 & \textbf{0.224} & 0.209 & 0.004 & 0.369 \\
 & Stage 1  & 0.002\tsm & 0.102\tdn & 0.101\tdn & 0.213\tdn & 0.173\tdn & 0.001\tsm & 0.353\tdn \\
 & Stage 2$^\dagger$ & \multicolumn{7}{c}{\textit{Generation Collapse  -  empty outputs}} \\
 & Stage 3  & 0.001\tup & \textbf{0.636}\tup & \textbf{0.636}\tup & 0.074\tup & \textbf{0.802}\tup & 0.001\tsm & \textbf{0.619}\tup \\
 & Stage 4  & 0.001\tsm & \textbf{0.637}\tsm & \textbf{0.637}\tsm & 0.074\tsm & \textbf{0.803}\tsm & \textbf{0.001}\tsm & \textbf{0.619}\tsm \\
\midrule
\multicolumn{9}{l}{\textit{Decoder-Only Models}} \\
\midrule
\multirow{5}{*}{\llama{} (1B)}
 & Baseline & 0.023 & 0.198 & 0.197 & 0.356 & 0.111 & 0.107 & 0.661 \\
 & Stage 1  & 0.010\tdn & 0.053\tdn & 0.052\tdn & 0.377\tup & 0.216\tup & 0.497\tup & 0.592\tdn \\
 & Stage 2  & 0.006\tdn & 0.044\tdn & 0.043\tdn & 0.424\tup & 0.089\tdn & \textbf{0.019}\tdn & 0.565\tdn \\
 & Stage 3  & 0.077\tup & \textbf{0.624}\tup & \textbf{0.622}\tup & \textbf{0.555}\tup & \textbf{0.783}\tup & 0.030\tup & 0.708\tup \\
 & Stage 4  & \textbf{0.079}\tsm & 0.611\tsm & 0.609\tsm & 0.540\tsm & 0.777\tsm & 0.031\tsm & \textbf{0.709}\tsm \\
\midrule
\multirow{5}{*}{\gemma{} (2B)}
 & Baseline & 0.016 & 0.088 & 0.086 & 0.267 & 0.490 & 0.004 & 0.624 \\
 & Stage 1  & 0.006\tdn & 0.021\tdn & 0.021\tdn & \textbf{0.494}\tup & 0.148\tdn & 0.003\tsm & 0.626\tsm \\
 & Stage 2  & 0.022\tup & 0.070\tup & 0.070\tup & 0.107\tdn & 0.691\tup & 0.001\tsm & 0.660\tup \\
 & Stage 3  & 0.153\tup & 0.718\tup & \textbf{0.717}\tup & 0.361\tup & \textbf{0.812}\tup & \textbf{0.000}\tdn & 0.780\tup \\
 & Stage 4  & \textbf{0.155}\tsm & \textbf{0.719}\tsm & \textbf{0.717}\tsm & 0.366\tsm & \textbf{0.812}\tsm & \textbf{0.000}\tsm & \textbf{0.781}\tsm \\
\midrule
\multirow{5}{*}{\mistral{} (7B)}
 & Baseline & 0.011 & 0.028 & 0.028 & 0.159 & 0.116 & 0.119 & 0.637 \\
 & Stage 1  & 0.014\tsm & 0.044\tup & 0.044\tup & \textbf{0.338}\tup & 0.349\tup & 0.148\tup & 0.629\tsm \\
 & Stage 2  & 0.014\tsm & 0.044\tsm & 0.044\tsm & 0.278\tdn & 0.390\tup & \textbf{0.021}\tdn & 0.624\tsm \\
 & Stage 3  & 0.069\tup & 0.393\tup & 0.392\tup & 0.245\tdn & 0.522\tup & 0.028\tsm & 0.699\tup \\
 & Stage 4  & \textbf{0.070}\tsm & \textbf{0.393}\tsm & \textbf{0.392}\tsm & 0.249\tsm & \textbf{0.531}\tsm & 0.028\tsm & \textbf{0.700}\tsm \\
\bottomrule
\end{tabular}}%
\end{table*}

\paragraph{Key observations.}
\begin{enumerate}[leftmargin=*, itemsep=2pt]
    \item \textbf{Stage~3 is the Hindi game-changer.} For Base, Hindi ROUGE-1 jumps from 0.481 to 0.691 (+0.210) - the largest single-stage improvement in our study. This confirms that bilingual SFT with Hindi examples is essential.

    \item \textbf{Cross-lingual citation transfer.} Stage~2 (English-only SFT) improves Hindi Citation F1 from 0.485 to 0.718 for Base, demonstrating that citation formatting is a language-agnostic structural skill that transfers cross-lingually.

    \item \textbf{Stage~1 is critical for small models.} Base Hindi BERTScore jumps from 0.221 to 0.351 (+0.130) after Stage~1, while Large shows minimal change (0.617 to 0.622). This confirms that multilingual adaptation primarily benefits models with limited pretraining Hindi exposure.

    \item \textbf{Base overtakes Large after Stage~3.} An interesting crossover: Base achieves Hindi ROUGE-1 of 0.691 vs.\ Large's 0.674, and Hindi Citation F1 of 0.812 vs.\ 0.798. The smaller model benefits more from bilingual SFT because it has more room to improve.

    \item \textbf{LLaMA-1B Hindi citation learning surpasses Mistral-7B.} Despite failing entirely on English citations, LLaMA-3.2-1B achieves Hindi Citation-F1 of 0.783 at Stage~3 - exceeding Mistral-7B's 0.522. The model appears to have allocated its limited citation learning capacity entirely to Hindi, the language that constituted 60\% of Stage~3 training examples. This language-selective learning is discussed further in Section~\ref{sec:llama_failure}.

    \item \textbf{Gemma-2-2B leads Hindi performance.} Gemma-2-2B achieves the strongest Hindi gains among decoder-only models: Hindi ROUGE-1 of 0.719 and Citation-F1 of 0.812 at Stage~4, comparable to the encoder-decoder models. This underscores that the progressive pipeline benefits decoder-only models as well, provided the model has sufficient capacity.
\end{enumerate}

\begin{figure*}[t]
\centering
\includegraphics[width=0.88\textwidth]{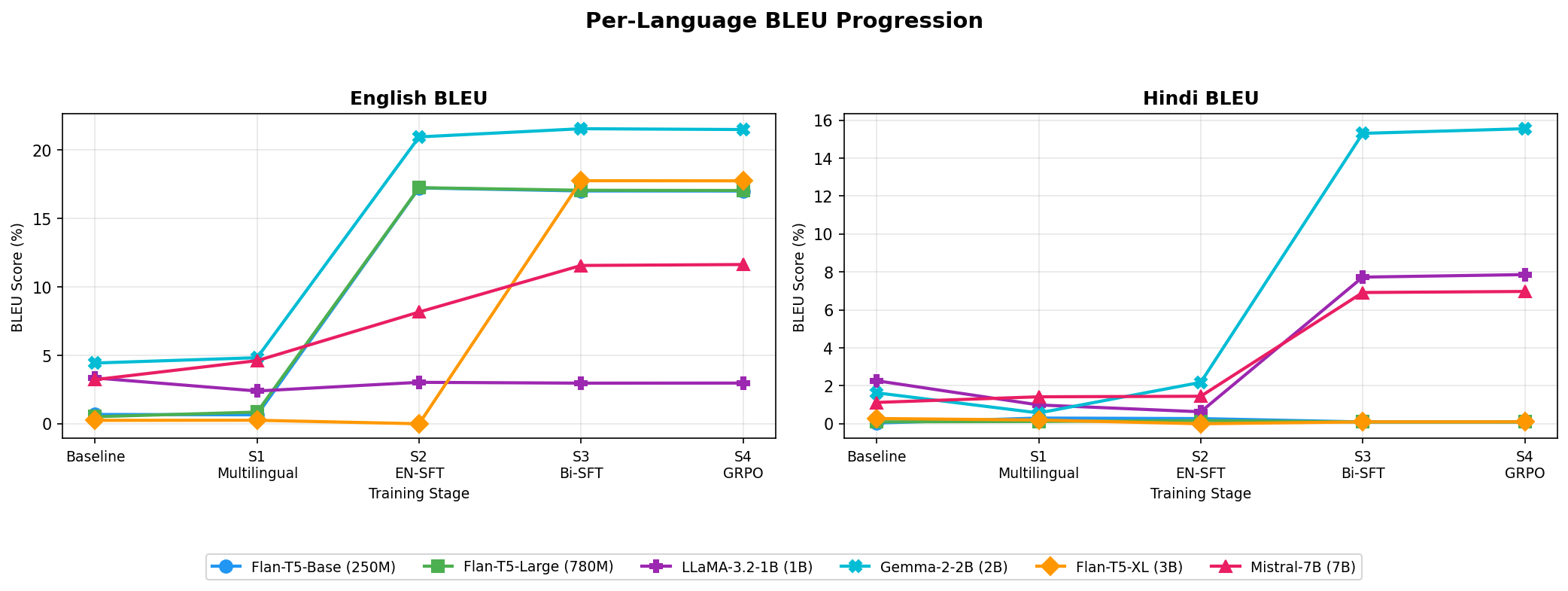}
\caption{Per-language BLEU progression across training stages. English BLEU (left) shows convergence of all models after Stage~2 SFT, while Hindi BLEU (right) remains near-zero for encoder-decoder models due to morphological variation (Section~\ref{sec:disc_hindi_bleu}). Gemma-2-2B achieves the highest Hindi BLEU (0.155) after Stage~3, followed by Mistral-7B (0.070).}
\label{fig:language_comparison}
\end{figure*}

\subsection{Flan-T5-XL Generation Collapse}
\label{sec:xl_collapse}

Flan-T5-XL (3B) exhibited \emph{generation collapse} after Stage~2 SFT: the model produced empty strings for all test examples, yielding zero scores across all metrics. However, the model \textbf{recovered fully in Stage~3} (Bilingual SFT), achieving performance comparable to Base and Large (overall BLEU 0.096, Citation F1 0.898, BERTScore 0.765).

\paragraph{Diagnosis.} The collapse is notable because the Stage~2 checkpoint appeared healthy by standard indicators: (i)~teacher-forced evaluation loss was 1.642 (comparable to Base at 1.916 and Large at 1.776), (ii)~model weights were intact (5.4\,GB across two safetensor shards), and (iii)~the generation configuration was correct (standard T5 defaults).

\paragraph{Root cause.} We attribute the collapse to the learning rate of $2 \times 10^{-5}$, which is too aggressive for a 3B encoder-decoder model. The successful models used proportionally lower rates: Base (250M) at $5 \times 10^{-5}$ and Large (780M) at $3 \times 10^{-5}$. At 3B parameters, the loss landscape is sharper, and the learning rate pushed the model into a degenerate local minimum where emitting the EOS token as the first generated token minimizes the autoregressive loss.

\paragraph{Recovery in Stage~3.} The bilingual SFT in Stage~3 effectively rescued the model from collapse. Despite building on the collapsed Stage~2 checkpoint, the continued training with a mixture of English and Hindi dialogue data pushed the model out of the degenerate minimum. After recovery, XL achieves English metrics (BLEU 0.177, Citation F1 0.980, BERTScore 0.891) comparable to Base and Large, and strong Hindi performance (ROUGE-1 0.636, Citation F1 0.802). This demonstrates the \textbf{resilience of continued training}: a collapsed checkpoint is not necessarily unrecoverable.

\paragraph{Implications.} This episode highlights a critical lesson: \textbf{teacher-forced evaluation loss is not a reliable proxy for generation quality}. A model can achieve reasonable teacher-forced loss while completely failing at autoregressive generation. Training pipelines for generative models should include periodic generation-time validation. This phenomenon is consistent with findings on neural text degeneration~\citep{holtzman2020curious} and unlikelihood training~\citep{welleck2020consistency}.

\subsection{LLaMA-3.2-1B: Language-Selective Citation Failure}
\label{sec:llama_failure}

LLaMA-3.2-1B exhibits a qualitatively distinct failure mode from Flan-T5-XL's generation collapse: \emph{language-selective citation failure}. From Stage~2 onward, the model generates zero citation markers in English (English Citation-F1 = 0.000, 0\% of responses contain citations), while simultaneously learning robust Hindi citation behavior by Stage~3 (Hindi Citation-F1 = 0.783, citations present in 94.2\% of responses).

\paragraph{Stage~1 hallucination explosion.} Unlike all encoder-decoder models, LLaMA-1B experiences a catastrophic hallucination spike after multilingual adaptation: overall hallucination rate rises from 13.5\% at baseline to 66.5\% after Stage~1 (English: 16.0\% $\to$ 81.0\%). This suggests that Stage~1 translation-style training is particularly disruptive for small decoder-only models.

\paragraph{Stage~2: Hallucination eliminated without citation format.} Stage~2 SFT reduces English hallucination from 81.0\% to 0.0\% - a dramatic recovery. However, English Citation-F1 collapses simultaneously from 0.287 (Stage~1) to 0.000, where it remains through Stages~3 and~4. The model learned a conservative generation strategy - avoid all specific claims and thus avoid hallucination - rather than grounding claims in cited passages. This demonstrates that \textbf{zero hallucination and zero citation are not contradictory}: a model can satisfy the former without the latter by generating only generic, non-committal text.

\paragraph{Stage~3: Hindi citation learning without English recovery.} The 60\%/40\% Hindi-English bilingual SFT in Stage~3 produces a striking asymmetry: Hindi Citation-F1 improves from 0.089 to 0.783 (surpassing Mistral-7B's 0.522), while English Citation-F1 remains 0.000. The model successfully learned citation format from Hindi training examples but neither generalised this behaviour to English nor recovered the English citation pathway lost in Stage~2. This is consistent with catastrophic forgetting in sequential fine-tuning~\citep{mccloskey1989catastrophic}: Stage~2 may have irreversibly overwritten English citation associations, and the Hindi-weighted Stage~3 did not re-establish them.

\paragraph{Qualitative evidence.} To verify that the zero English Citation-F1 is not a metric format mismatch (e.g., model generating \texttt{(1)} instead of \texttt{[1]}), we ran direct inference on the Stage~3 checkpoint against the citation test set. English outputs are citation-free and show repetition degeneration:

\begin{quote}
\small\textit{``No, you cannot bring your own liquor at Pizza Hut Cherry Hinton. Do you have any other questions? Would you like to make a reservation? If so, for what time and day? How many people will be dining? How many people? How many people?''}
\end{quote}

\noindent The expected output was: \textit{``As noted in [1], No, outside beverages are not allowed.''}. No citation marker appears in any English output across tested samples. Hindi outputs, by contrast, correctly produce citation markers embedded in fluent responses (e.g., \texttt{[2]} followed by citation-grounded Hindi text, translating to \textit{``Based on [2], I have booked you a taxi\ldots''}~\citep{li2024opera}). This confirms the citation failure is behavioural - the model architecture is capable of producing \texttt{[N]} markers but does not do so for English.

\paragraph{Stage~4 GRPO: No recovery.} GRPO training changes all metrics by at most $\pm$0.008 (Table~\ref{tab:grpo_delta}), confirming that RL alignment cannot recover citation behaviour that SFT failed to instil.

\paragraph{Implications.} LLaMA-3.2-1B's behaviour demonstrates two separable properties: (i)~hallucination elimination and citation format learning are independent objectives - a model can achieve 0\% hallucination without generating any citation markers; and (ii)~catastrophic forgetting of citation format can occur even within a carefully staged pipeline. Importantly, Gemma-2-2B (a 2B decoder-only model) achieves Citation-F1 of 0.903 overall and 0.980 on English from Stage~2 onward - demonstrating that citation format learning is achievable in decoder-only models at this parameter scale. This confirms that LLaMA-3.2-1B's citation failure is a \textbf{model-specific anomaly} rather than a general architectural limitation of decoder-only models, plausibly driven by its smaller effective capacity, distinct pretraining corpus, or instruction-tuning protocol.

\subsection{Stage 4 (GRPO) Effectiveness}
\label{sec:grpo_results}

A central question of our study is whether GRPO alignment provides measurable improvement over SFT. Table~\ref{tab:grpo_delta} presents the Stage~3 $\rightarrow$ Stage~4 deltas.

\begin{table}[t]
\centering
\caption{Change from Stage~3 to Stage~4 (GRPO). Deltas are negligible for encoder-decoder models. Mistral shows marginal gains in Citation F1 and FactScore.}
\label{tab:grpo_delta}
\small
\begin{tabular}{@{}llrr@{}}
\toprule
& \textbf{Metric} & \textbf{Stage~3} & \textbf{$\Delta$} \\
\midrule
\multicolumn{4}{l}{\textit{Encoder-Decoder}} \\
\midrule
\multirow{4}{*}{Base}
 & Cit-F1 & 0.902 & 0.000 \\
 & Halluc. & 0.000 & 0.000 \\
 & BERTScore & 0.766 & 0.000 \\
 & FactScore & 0.096 & $+$0.002 \\
\cmidrule{2-4}
\multirow{4}{*}{Large}
 & Cit-F1 & 0.896 & $-$0.001 \\
 & Halluc. & 0.000 & 0.000 \\
 & BERTScore & 0.762 & 0.000 \\
 & FactScore & 0.084 & 0.000 \\
\cmidrule{2-4}
\multirow{4}{*}{XL}
 & Cit-F1 & 0.898 & 0.000 \\
 & Halluc. & 0.000 & 0.000 \\
 & BERTScore & 0.765 & 0.000 \\
 & FactScore & 0.095 & 0.000 \\
\midrule
\multicolumn{4}{l}{\textit{Decoder-Only}} \\
\midrule
\multirow{4}{*}{LLaMA}
 & Cit-F1 & 0.362 & $-$0.003 \\
 & Halluc. & 0.014 & $\ $0.000 \\
 & BERTScore & 0.774 & $+$0.001 \\
 & FactScore & 0.401 & $-$0.008 \\
\cmidrule{2-4}
\multirow{4}{*}{Gemma}
 & Cit-F1 & 0.903 & $\ $0.000 \\
 & Halluc. & 0.000 & 0.000 \\
 & BERTScore & 0.845 & $\ $0.000 \\
 & FactScore & 0.226 & $+$0.003 \\
\cmidrule{2-4}
\multirow{4}{*}{Mistral}
 & Cit-F1 & 0.768 & $+$0.004 \\
 & Halluc. & 0.014 & 0.000 \\
 & BERTScore & 0.786 & $+$0.001 \\
 & FactScore & 0.151 & $+$0.004 \\
\bottomrule
\end{tabular}
\end{table}

\begin{figure*}[t]
\centering
\begin{minipage}[t]{0.48\textwidth}
    \centering
    \includegraphics[width=\textwidth]{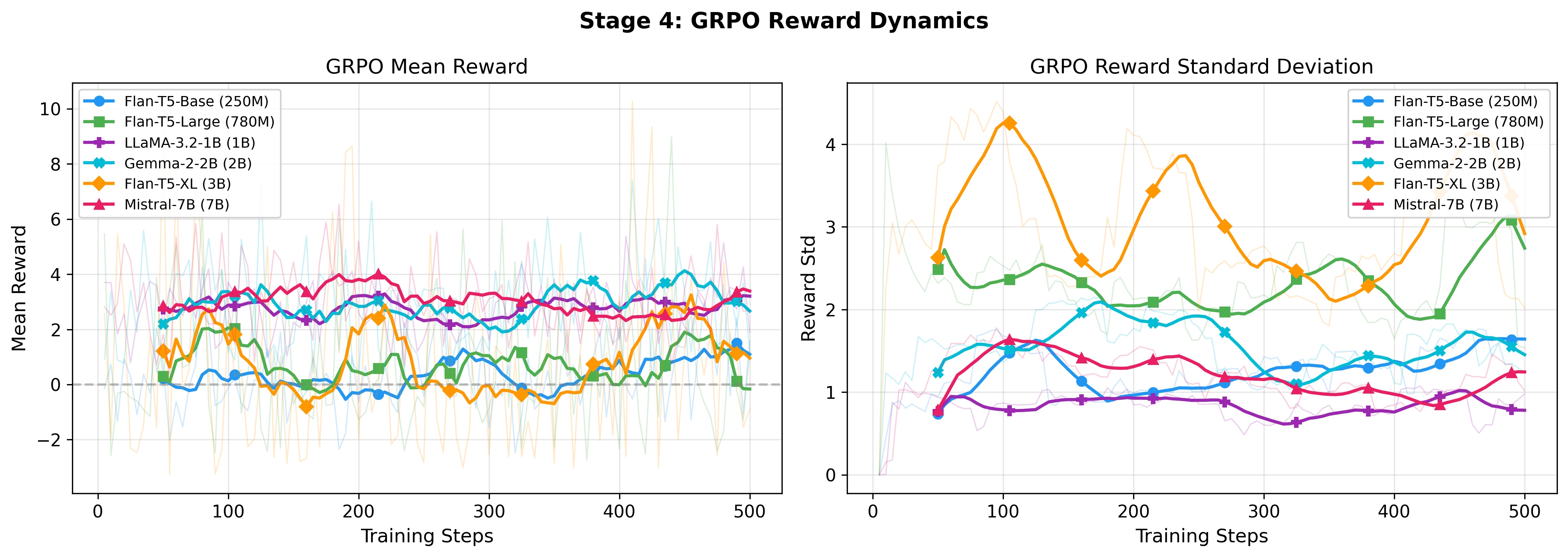}
    \captionof{figure}{GRPO reward dynamics during Stage~4. Left: mean reward with smoothed curves showing Mistral-7B consistently achieving the highest reward. Right: reward standard deviation, indicating training stability.}
    \label{fig:grpo_reward_dynamics}
\end{minipage}\hfill
\begin{minipage}[t]{0.48\textwidth}
    \centering
    \includegraphics[width=\textwidth]{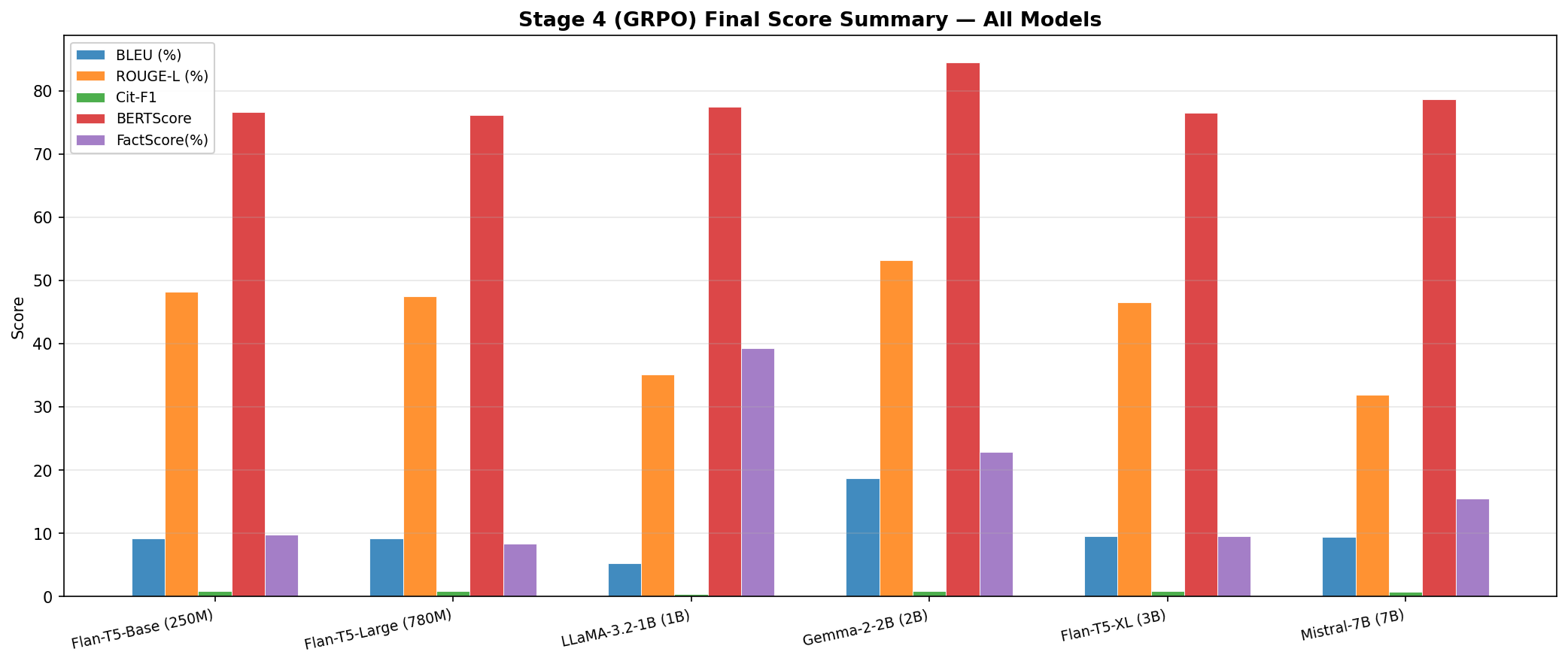}
    \captionof{figure}{Stage~4 (GRPO) final score summary across all six models, showing five metrics side by side. Gemma-2-2B and Mistral-7B lead in most metrics after GRPO alignment.}
    \label{fig:stage4_final_summary}
\end{minipage}
\end{figure*}

GRPO adds essentially zero improvement across all metrics for both models. We analyze the GRPO training dynamics in Table~\ref{tab:grpo_rewards} to understand why. The full reward trajectory is visualized in Figure~\ref{fig:grpo_reward_dynamics}, with reward distributions in Figure~\ref{fig:grpo_reward_distribution} and best-vs-final comparison in Figure~\ref{fig:grpo_reward_comparison} (see Appendix).

\begin{table}[t]
\centering
\caption{GRPO reward trajectory during Stage~4 training. Encoder-decoder and decoder-only models (LLaMA, Mistral) show reward decline from best to final, suggesting training instability. Gemma-2-2B is the exception: its best reward equals its final reward, indicating reward improvement continued to the last step.}
\label{tab:grpo_rewards}
\small
\begin{tabular}{@{}llrrr@{}}
\toprule
& \textbf{Model} & \textbf{Best} & \textbf{Final} & \textbf{$\Delta$} \\
\midrule
\multirow{3}{*}{\rotatebox[origin=c]{90}{\scriptsize Enc-Dec}}
 & Base (250M)  & 1.040 & 0.648 & $-$0.391\tdn \\
 & Large (780M) & 0.868 & 0.157 & $-$0.711\tdn \\
 & XL (3B)      & 1.782 & 0.482 & $-$1.300\tdn \\
\midrule
\multirow{3}{*}{\rotatebox[origin=c]{90}{\scriptsize Dec-Only}}
 & LLaMA (1B)   & 3.08  & 2.51  & $-$0.57\tdn \\
 & Gemma (2B)   & 3.40  & 3.40  & $\ $0.00 \\
 & Mistral (7B) & 2.889 & 2.433 & $-$0.457\tdn \\
\bottomrule
\end{tabular}
\end{table}

\paragraph{Analysis.} The reward declines from best to final for most models (Figure~\ref{fig:grpo_reward_comparison}), suggesting that GRPO training is unstable in this setting. The exception is Gemma-2-2B, which achieves the highest absolute reward (3.40) with zero decline - its best and final rewards are identical, indicating improvement continued to the last training step. Among models that show reward regression, Mistral-7B achieves the highest absolute reward (2.889) and the smallest decline ($-$0.457). We discuss possible causes in Section~\ref{sec:discussion}.

\subsection{Training Efficiency}
\label{sec:training_efficiency}

Table~\ref{tab:training_summary} summarizes training steps across all stages. The complete training loss curves are presented in Figures~\ref{fig:loss_grid} and~\ref{fig:val_loss} (see Appendix).

\begin{table}[t]
\centering
\caption{Training summary showing total steps and best validation loss per stage. Mistral-7B consistently achieves the lowest loss across all stages.}
\label{tab:training_summary}
\small
\begin{tabular}{@{}lllrr@{}}
\toprule
& \textbf{Model} & \textbf{Stage} & \textbf{Steps} & \textbf{Best Loss} \\
\midrule
\multirow{9}{*}{\rotatebox[origin=c]{90}{\scriptsize Encoder-Decoder}}
 & \multirow{3}{*}{Base}  & Stage 1 & 3,906  & 1.552 \\
 &                        & Stage 2 & 7,500  & 1.916 \\
 &                        & Stage 3 & 4,000  & 0.945 \\
\cmidrule{2-5}
 & \multirow{3}{*}{Large} & Stage 1 & 3,906  & 1.439 \\
 &                        & Stage 2 & 8,973  & 1.776 \\
 &                        & Stage 3 & 4,000  & 0.912 \\
\cmidrule{2-5}
 & \multirow{3}{*}{XL}    & Stage 1 & 7,812  & 1.384 \\
 &                        & Stage 2 & 14,000 & 1.642 \\
 &                        & Stage 3 & 6,000  & 0.895 \\
\midrule
\multirow{9}{*}{\rotatebox[origin=c]{90}{\scriptsize Decoder-Only}}
 & \multirow{3}{*}{LLaMA} & Stage 1 & 7,812  & 1.408 \\
 &                        & Stage 2 & 7,500  & 1.930 \\
 &                        & Stage 3 & 12,000 & 0.912 \\
\cmidrule{2-5}
 & \multirow{3}{*}{Gemma} & Stage 1 & 7,812  & 1.458 \\
 &                        & Stage 2 & 6,000  & 1.104 \\
 &                        & Stage 3 & 9,500  & 0.651 \\
\cmidrule{2-5}
 & \multirow{3}{*}{Mistral} & Stage 1 & 7,812  & 0.742 \\
 &                          & Stage 2 & 7,000  & 0.927 \\
 &                          & Stage 3 & 9,500  & 0.303 \\
\bottomrule
\end{tabular}
\end{table}

Mistral-7B achieves substantially lower loss across all stages (e.g., Stage~3 loss of 0.303 vs.\ XL's 0.895), suggesting that the decoder-only 7B model has significantly more capacity for this task. The Stage~2 $\rightarrow$ Stage~3 loss drop is consistent across all models, confirming that bilingual SFT improves the model's fit to the dialogue distribution.

\subsection{Explainability Analysis}
\label{sec:explainability}

We conducted three complementary explainability analyses on representative test examples: (i)~attention alignment, measuring cross-attention focus on cited knowledge tokens; (ii)~gradient saliency, quantifying input token attribution; and (iii)~occlusion sensitivity, testing whether citations are causally grounded in their sources.

\paragraph{Attention alignment.}
For encoder-decoder models (Flan-T5 Base, Large, XL), we measure the mean cross-attention weight assigned to cited knowledge tokens during generation. Decoder-only models (Gemma, LLaMA, Mistral) lack a distinct cross-attention mechanism and yield 0.000 by construction (see Section~\ref{sec:disc_interpretability}). Table~\ref{tab:attention_alignment} summarises per-model, per-stage alignment scores.

\begin{table}[h]
\centering
\caption{Mean cross-attention alignment with cited knowledge tokens (encoder-decoder models only). Decoder-only models show 0.000 by architecture. Stage~2 collapse for XL is marked with $\dagger$.}
\label{tab:attention_alignment}
\small
\begin{tabular}{@{}lccccc@{}}
\toprule
\textbf{Model} & \textbf{Base} & \textbf{S1} & \textbf{S2} & \textbf{S3} & \textbf{S4} \\
\midrule
Flan-T5-Base  & 0.017 & 0.022 & 0.035 & 0.019 & 0.019 \\
Flan-T5-Large & 0.030 & 0.032 & 0.037 & 0.022 & 0.022 \\
Flan-T5-XL    & 0.003 & 0.004 & 0.000$^\dagger$ & 0.026 & 0.026 \\
\midrule
Gemma-2-2B    & \multicolumn{5}{c}{\textit{N/A (decoder-only)}} \\
LLaMA-3.2-1B  & \multicolumn{5}{c}{\textit{N/A (decoder-only)}} \\
Mistral-7B    & \multicolumn{5}{c}{\textit{N/A (decoder-only)}} \\
\bottomrule
\end{tabular}
\end{table}

\begin{figure}[h]
\centering
\includegraphics[width=0.9\columnwidth]{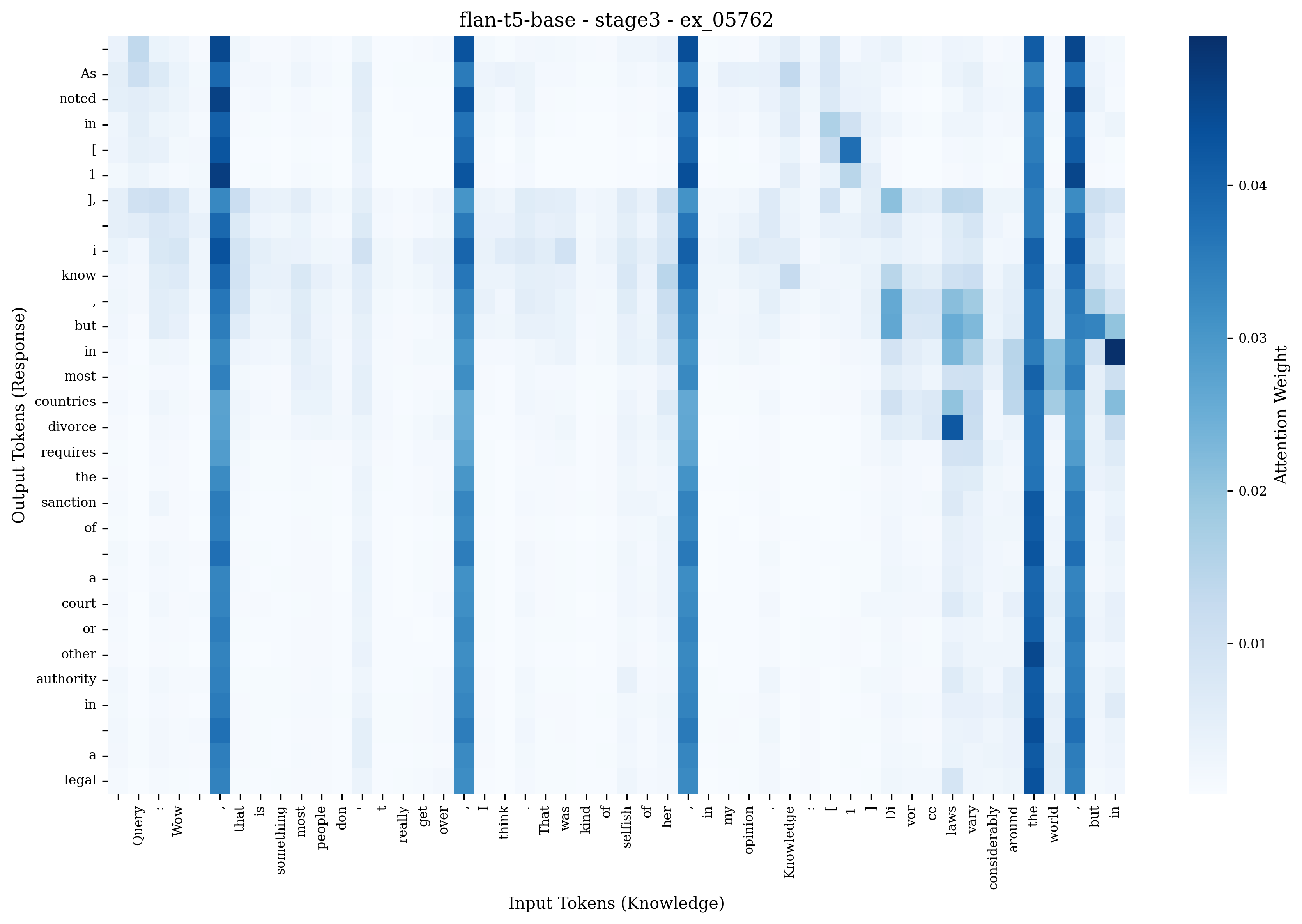}
\caption{Representative cross-attention heatmap for Flan-T5-Base (Stage~3). Output tokens (y-axis) attend to input knowledge tokens (x-axis). Brighter cells indicate higher attention weight. Citation-relevant tokens receive disproportionately high attention.}
\label{fig:attention_heatmap}
\end{figure}

All three Flan-T5 variants show peak alignment at Stage~2 (English SFT), the stage with the most dramatic citation metric improvement. Flan-T5-XL's Stage~2 alignment of 0.000 corroborates the generation collapse detected by evaluation metrics. After recovery at Stage~3, XL's alignment (0.026) exceeds Base (0.019) and Large (0.022), consistent with its larger capacity. A representative cross-attention heatmap for Flan-T5-Base at Stage~3 is shown in Figure~\ref{fig:attention_heatmap}.

\paragraph{Gradient saliency.}
For encoder-decoder models, we computed gradient-based token saliency at the input embedding layer and measured saliency entropy (spread of attribution across tokens) and concentration (fraction attributed to top-5 tokens). Decoder-only models have empty saliency records due to architecture differences in gradient flow through causal attention.

As shown in Table~\ref{tab:saliency}, entropy \emph{increases} through training for all three Flan-T5 models, while concentration \emph{decreases}. This indicates that well-trained models distribute attention more broadly across the knowledge passage rather than focusing rigidly on a small set of tokens. Flan-T5-XL Stage~2 shows NaN saliency (entropy and concentration undefined), providing independent confirmation of the generation collapse where no meaningful gradient signal flows from the output.

\begin{table}[h]
\centering
\caption{Gradient saliency entropy and concentration for Flan-T5 encoder-decoder models across training stages. Higher entropy = more diffuse attribution; lower concentration = less dominated by top tokens. XL Stage~2 NaN confirms generation collapse.}
\label{tab:saliency}
\small
\begin{tabular}{@{}lcccccc@{}}
\toprule
& \multicolumn{2}{c}{\textbf{Baseline}} & \multicolumn{2}{c}{\textbf{Stage~2}} & \multicolumn{2}{c}{\textbf{Stage~4}} \\
\cmidrule(lr){2-3}\cmidrule(lr){4-5}\cmidrule(lr){6-7}
\textbf{Model} & \textbf{Entr.} & \textbf{Conc.} & \textbf{Entr.} & \textbf{Conc.} & \textbf{Entr.} & \textbf{Conc.} \\
\midrule
Base  & 4.36 & 0.378 & 4.44 & 0.343 & 4.50 & 0.326 \\
Large & 4.45 & 0.327 & 4.48 & 0.322 & 4.49 & 0.320 \\
XL    & 4.38 & 0.360 & NaN  & NaN   & 4.53 & 0.306 \\
\bottomrule
\end{tabular}
\end{table}

\paragraph{Occlusion sensitivity.}
We measure causal grounding as the fraction of citation markers that \emph{disappear} when the corresponding source passage is removed from the input. A high score indicates the citation is genuinely conditioned on the source. Table~\ref{tab:occlusion} reports causal grounding at Stage~3 (best-performing stage) for all models.

\begin{table}[h]
\centering
\caption{Occlusion causal grounding: fraction of citations that disappear when source is removed. Higher = stronger causal grounding. Both Mistral-7B and Gemma-2-2B drop to 0.000 post-training, indicating decoder-only models learn citation format without source-conditioned grounding.}
\label{tab:occlusion}
\small
\begin{tabular}{@{}lcccc@{}}
\toprule
\textbf{Model} & \textbf{Baseline} & \textbf{Stage~1} & \textbf{Stage~3} & \textbf{Stage~4} \\
\midrule
Flan-T5-Base  & 0.647 & 0.691 & 0.889 & 0.889 \\
Flan-T5-Large & 0.656 & 0.630 & 0.909 & 0.909 \\
Flan-T5-XL    & 0.960 & 0.964 & 0.808 & 0.808 \\
LLaMA-3.2-1B  & 0.696 & 0.544 & 0.722 & 0.632 \\
Gemma-2-2B$^\ddagger$    & 0.733 & 0.000 & 0.000 & 0.000 \\
Mistral-7B$^\dagger$    & 0.767 & 0.797 & 0.000 & 0.000 \\
\bottomrule
\end{tabular}
\par\vspace{3pt}
\begin{minipage}{\columnwidth}
\scriptsize
$^\dagger$Mistral Stage~3--4: citations generated but not causally grounded.\\[1pt]
$^\ddagger$Gemma Stage~1--4: citation generation suppressed on these 100 explainability examples post-training (Stage~1 subset CF1=0.186, Stages~3--4 = 0.000); overall test-set Citation-F1 is substantially higher.
\end{minipage}
\end{table}

Flan-T5-Base and Large show substantial improvement in causal grounding through training (0.647/0.656 at baseline to 0.889/0.909 at Stage~3), indicating that SFT trains the model to produce citations that are genuinely source-dependent. Flan-T5-XL starts with very high baseline grounding (0.960) - consistent with its high baseline Citation-F1 (0.610) - but declines slightly at Stage~3 (0.808) after its collapse-and-recovery. LLaMA-3.2-1B's grounding recovers at Stage~3 (0.722) despite its English citation failure, plausibly because Hindi citations remain source-dependent.

A striking architecture-level pattern emerges: \textbf{both decoder-only models collapse to 0.000 causal grounding after fine-tuning}. Mistral-7B drops from 0.767 (baseline) to 0.000 at Stage~3--4 despite Citation-F1 of 0.772, and Gemma-2-2B drops from 0.733 (baseline) to 0.000 from Stage~1 onward despite Citation-F1 of 0.903 at Stage~3. In both cases, the model learns to produce citation \emph{format} (marker tokens) without grounding the cited content in the provided knowledge source. This decoder-only format-grounding dissociation contrasts sharply with encoder-decoder models, where cross-attention provides a structural mechanism for source conditioning that persists through fine-tuning.


\section{Discussion}
\label{sec:discussion}

We organize our discussion around seven key findings and their broader implications for knowledge-grounded dialogue research.

\subsection{Citation-Grounded SFT as Anti-Hallucination}
\label{sec:disc_hallucination}

Under automatic NLI-based evaluation, hallucination rate drops to 0.0\% after Stage~2 for encoder-decoder models (Flan-T5 Base and Large) and remains there through Stages~3 and~4. For Mistral-7B (decoder-only), hallucination drops from 0.078 to 0.010 after Stage~2, stabilizing at 0.014 after Stage~3. This architecture-dependent pattern suggests that encoder-decoder models, with their explicit cross-attention over knowledge passages, may be more amenable to hallucination reduction through citation-grounded SFT. We note that these results are based on automatic metrics; human evaluation may reveal a different picture and remains an important avenue for future work.

The likely mechanism is that when every training example contains properly cited responses of the form ``According to [1], \ldots'', the model learns to associate claims with numbered references. At inference time, this encourages the model to ground claims in provided passages rather than generating unsupported content. Whether this transfers fully to real-world deployment conditions warrants further study.

\paragraph{LLaMA-1B counterexample: hallucination without citation grounding.} LLaMA-3.2-1B's results challenge the assumption that citation-grounded SFT is the \emph{mechanism} by which hallucination is reduced. From Stage~2 onward, LLaMA-1B achieves 0.0\% English hallucination - yet its English Citation-F1 is simultaneously 0.000 (no citation markers generated). The model suppresses hallucination not by citing passages but by adopting a conservative, non-committal generation strategy that avoids specific factual claims altogether. This demonstrates that \textbf{zero hallucination and zero citation grounding are not contradictory}, and that the causal path from SFT to hallucination reduction may be model-specific.

These observations suggest that citation-grounded SFT may be a promising direction for reducing hallucination, potentially complementing post-hoc detection approaches. We encourage future work to validate these findings with human evaluation across diverse domains.

\subsection{SFT Sufficiency for Structured Tasks}
\label{sec:disc_sft_sufficiency}

The marginal gains observed after Stage~4 GRPO (Section~\ref{sec:grpo_results}) invite a broader question: \emph{under what conditions does RL alignment provide meaningful benefit over well-tuned SFT?}

We offer one possible interpretation: citation-grounded dialogue has a well-defined output format with relatively clear quality criteria. SFT directly optimises for this format by maximising likelihood of reference outputs. When the reference outputs already satisfy quality criteria, the RL signal may have limited additional structure to exploit. We note, however, that our experimental configuration represents one point in a large hyperparameter space, and we do not claim this as a general conclusion.

We identify four factors that may have limited GRPO's contribution in our specific setting:

\begin{enumerate}[leftmargin=*, itemsep=2pt]
    \item \textbf{Possible reward saturation.} Citation quality and hallucination metrics were near-optimal after SFT, which may have reduced the reward signal available for RL to exploit.
    \item \textbf{KL penalty strength.} The coefficient $\beta = 0.04$ constrains the policy to stay close to the SFT checkpoint; a lower $\beta$ may allow more beneficial exploration.
    \item \textbf{Reward signal granularity.} The citation component is largely binary, which may provide limited gradient signal when the model already cites consistently.
    \item \textbf{Training budget.} At 500 steps with group size 4, only 2,000 candidate responses are explored; longer training may yield different outcomes.
\end{enumerate}

Notably, both Gemma-2-2B (3.40) and Mistral-7B (2.889) achieve substantially higher absolute rewards than encoder-decoder models (0.48--1.78), suggesting that RL behaviour may be scale- and architecture-dependent. Gemma-2-2B's zero reward decline is particularly noteworthy: it may indicate that a longer GRPO training budget would yield further improvement for this model. Systematic hyperparameter exploration - varying $\beta$, training steps, group size, and reward design - remains an important direction for future work.

\subsection{The Hindi BLEU Problem}
\label{sec:disc_hindi_bleu}

Hindi BLEU remains near-zero for encoder-decoder models ($\leq$0.003) across all stages, even when ROUGE-1 (up to 0.691) and BERTScore (up to 0.624) indicate strong Hindi generation quality. Decoder-only models achieve somewhat higher but still modest BLEU values (LLaMA-3.2-1B: 0.079, Gemma-2-2B: 0.155), yet these too substantially underestimate quality relative to ROUGE-1 (0.624, 0.719) and BERTScore (0.709, 0.781) respectively. This is not a model failure but a \textbf{metric limitation} (see Figure~\ref{fig:language_comparison}).

BLEU~\citep{papineni2002bleu} measures exact $n$-gram precision, which is inherently biased against morphologically rich languages. Hindi exhibits: (a)~productive morphological inflection, (b)~relatively free word order, and (c)~multiple valid surface realizations. This observation is consistent with \citet{post2018clarity}. For Hindi evaluation, we recommend \textbf{BERTScore as the primary metric}, with ROUGE-1 as a secondary signal.

\subsection{Contextualising Absolute Performance}
\label{sec:disc_context}

To place our results in context, we compare against FaithDial~\citep{rashkin2021faithdial}, the most closely related benchmark for faithful knowledge-grounded dialogue evaluation. The best-performing FaithDial model (T5-CTRL trained on FaithDial+WoW) reports BLEU of 0.115 and BERTScore of 0.835. Our Stage~3 encoder-decoder models achieve BERTScore of 0.762--0.891, matching or exceeding FaithDial on semantic quality, while Mistral-7B reaches 0.787. Our BLEU (0.092--0.095) is slightly below FaithDial's 0.115, which we attribute partly to the structured citation constraint: our responses contain explicit citation markers (\texttt{[1]}, \texttt{[2]}) that are absent from free-form FaithDial references, reducing exact $n$-gram overlap by design. Importantly, FaithDial does not evaluate citation accuracy or attribution - tasks that our pipeline additionally addresses. Direct cross-paper comparison should be interpreted cautiously given differences in test sets, knowledge sources, and evaluation protocols. We treat BERTScore as the primary semantic quality metric and Citation-F1 with hallucination rate as the primary grounding metrics, with BLEU serving as a secondary indicator.

\subsection{Scaling Behavior and Architecture Effects}
\label{sec:disc_scaling}

Our results reveal nuanced interactions between model size, architecture, and training stage:

\paragraph{English performance saturates early.} After Stage~2, both Base (250M) and Large (780M) achieve identical English metrics (BLEU 0.172, BERTScore 0.889, Citation F1 0.980). This suggests that for constrained, well-defined tasks, model capacity beyond 250M provides no English benefit. This has significant deployment implications: a 250M model is $3\times$ cheaper to serve than 780M with no quality loss.

\paragraph{Hindi benefits from scale at baseline.} Large starts with Hindi BERTScore of 0.617 vs.\ Base's 0.221 - a $2.8\times$ gap attributable to more Hindi data in the larger pretraining corpus. However, this gap narrows through training: after Stage~3, Base reaches 0.624 vs.\ Large's 0.615, effectively closing the gap. This demonstrates that our progressive pipeline can compensate for limited pretraining Hindi exposure.

\paragraph{Generation collapse is recoverable.} Flan-T5-XL (3B, encoder-decoder) collapsed at Stage~2 under a learning rate of $2 \times 10^{-5}$, but recovered fully at Stage~3 (Bilingual SFT), achieving performance comparable to Base and Large. This demonstrates that continued training can rescue a collapsed checkpoint, and that encoder-decoder models at scale require more careful learning rate tuning - Mistral-7B trains with $1 \times 10^{-5}$ without collapse.

\paragraph{Decoder-only vs.\ encoder-decoder trade-offs.} Mistral-7B achieves higher absolute BLEU than encoder-decoder models at baseline (0.023 vs.\ 0.003--0.004) and maintains a lead in FactScore throughout training (0.155 at Stage~4 vs.\ 0.095--0.098 for Flan-T5). However, Mistral does not eliminate hallucination entirely (0.014 at Stage~3/4 vs.\ 0.000 for Flan-T5 Base/Large), and its Hindi Citation F1 (0.531) lags behind the encoder-decoder models (0.797--0.803). This suggests that encoder-decoder architectures are more effective at learning the citation grounding constraint.

\paragraph{Heterogeneous decoder-only behavior.} The decoder-only models in our study show strikingly different outcomes. Mistral-7B (7B) successfully learns English citation format (Citation-F1 0.979 at Stage~4), while LLaMA-3.2-1B (1B) fails entirely (Citation-F1 0.000 from Stage~2 onward). Gemma-2-2B (2B), however, achieves Citation-F1 of 0.980 on English and 0.812 on Hindi by Stage~3 - demonstrating that the failure is LLaMA-specific rather than a scale threshold effect: a 2B decoder-only model \emph{can} learn bilingual citation format under our pipeline. This indicates that \textbf{citation learning failure is not an architectural property of decoder-only models} and is not simply a function of parameter count; it is plausibly driven by LLaMA-3.2-1B's specific pretraining corpus composition, instruction-tuning protocol, or optimizer state.

\paragraph{Larger models hallucinate more before grounding.} Large (780M) has a baseline hallucination rate of 0.078, Mistral-7B 0.078, vs.\ Base's 0.005. Larger models generate more fluent, confident text, which manifests as more plausible-sounding but unsupported claims. This makes knowledge grounding training \emph{more important} for larger models, not less.

\subsection{Citation Phrase Salience and Language Module Strength}
\label{sec:disc_salience}

Output inspection of LLaMA-3.2-1B Stage~3 (Section~\ref{sec:llama_failure}) enables a hypothesis about why citation learning succeeds in Hindi but fails in English. We propose that citation learnability is jointly determined by two factors:

\paragraph{Citation phrase salience.} Hindi citation phrases such as ``\textit{[1]~\hi{में कहा गया है}}'' (it is said in [1]) or ``\textit{[2]~\hi{के आधार पर}}'' (based on [2]) are formulaic, high-salience constructions - the citation marker and its framing phrase form an inseparable grammatical unit that stands out as a distinctive pattern during training. English citation phrases (``According to [1]'', ``As noted in [1]'') are semantically richer but also more varied and closer to non-cited paraphrase phrasing, making them lower-salience patterns.

\paragraph{Pretrained language module strength.} LLaMA-3.2-1B's English baseline BERTScore (0.842) substantially exceeds its Hindi BERTScore (0.661), reflecting stronger English pretraining. We hypothesize that strong pretrained English fluency creates a resistance to the citation pattern: during Stage~2 SFT (training loss 1.930 $>$ Stage~1 loss 1.408), the gradient signal from citation learning was insufficient to overcome the inertia of English fluency representations. Hindi, where the pretrained module is weaker, offered less resistance to learning the new citation pattern during Stage~3.

\paragraph{Implication.} If this hypothesis holds, it suggests a design principle for citation format selection: citation markers should be maximally distinctive from natural-language paraphrase patterns, especially when fine-tuning strong pretrained models. Alternatives such as \texttt{\textless{}cite:1\textgreater{}} or \texttt{[SOURCE:1]} may be more learnable for high-resource languages than the conventional \texttt{[1]} notation. We acknowledge this remains a hypothesis requiring controlled ablation studies (e.g., varying citation format style while holding all else constant) to verify.

\subsection{Interpretability of Citation Behaviour}
\label{sec:disc_interpretability}

The explainability analyses (Section~\ref{sec:explainability}) reveal qualitatively distinct mechanisms of citation grounding across architectures.

\paragraph{Encoder-decoder cross-attention as citation grounding.}
Flan-T5 models possess a cross-attention sublayer that directly attends to encoder representations of the knowledge passage during decoding. Our attention alignment scores (Table~\ref{tab:attention_alignment}) confirm that cited tokens receive disproportionate cross-attention weight: mean alignment increases from 0.017 at baseline to a peak of 0.035--0.037 at Stage~2, the stage where Citation-F1 shows its most dramatic jump. The Stage~2 SFT teaches the decoder to route attention toward cited passage segments at the moment of generating citation markers - a direct, mechanistically interpretable grounding signal.

\paragraph{Decoder-only models: absence of cross-attention.}
Decoder-only architectures (Gemma-2-2B, LLaMA-3.2-1B, Mistral-7B) process the full prompt (query + knowledge passages) as a single causal sequence. There is no dedicated cross-attention sublayer; knowledge is accessed through the same self-attention mechanism as all other context. Cross-attention alignment is therefore architecturally 0.000 for these models - a measurement constraint, not a failure. These models must learn to retrieve and cite knowledge using self-attention alone, a fundamentally different mechanism that requires the model to internally route long-range attention from the response token to the relevant source token positions.

\paragraph{Saliency as a training quality signal.}
The monotonic decrease in saliency concentration across training stages (Table~\ref{tab:saliency}) suggests that well-trained models distribute credit more evenly across input tokens, consistent with broader contextual integration. Importantly, Flan-T5-XL's Stage~2 NaN saliency provides an \emph{independent} confirmation of generation collapse that complements the zero-score evaluation result: no gradient signal flows when the model generates empty outputs, making saliency undefined.

\paragraph{Decoder-only format-grounding dissociation.}
A consistent architecture-level pattern emerges from occlusion analysis: \emph{both} decoder-only models exhibit 0.000 causal grounding after fine-tuning. Mistral-7B drops from 0.767 to 0.000 at Stage~3--4 despite Citation-F1 of 0.772, and Gemma-2-2B drops from 0.733 to 0.000 from Stage~1 onward despite Citation-F1 of 0.903 at Stage~3 (Table~\ref{tab:occlusion}). This dissociation - citation format present but causal grounding absent - indicates that decoder-only models learn to produce the syntactic pattern of citation (inserting \texttt{[N]} markers at contextually appropriate positions) without the semantic grounding of citation content (claims genuinely conditioned on the removed source). The model may generate correct citation \emph{numbers} based on positional or conversational context rather than genuinely consulting source content. That this pattern holds across two architecturally distinct decoder-only models (Mistral's sliding-window attention vs.\ Gemma's grouped-query attention) suggests it is a general property of causal language models fine-tuned on citation tasks, not an artefact of a single architecture. This finding strengthens the argument that Citation-F1 alone is an insufficient grounding metric and must be complemented by occlusion-based causal grounding tests.

\subsection{Limitations}
\label{sec:limitations}

We acknowledge several limitations:

\begin{enumerate}[leftmargin=*, itemsep=2pt]
    \item \textbf{Hindi data quality.} Hindi examples are machine-translated, which may not fully reflect natural conversational patterns.
    \item \textbf{Single-reference evaluation.} All automatic metrics compare against a single reference response.
    \item \textbf{GRPO exploration.} Comprehensive hyperparameter search might reveal configurations where GRPO provides measurable improvement.
    \item \textbf{Human evaluation.} This study relies entirely on automatic metrics.
    \item \textbf{Explainability scope.} Attention alignment analysis is limited to encoder-decoder models with cross-attention; comparable interpretability tools for decoder-only models (e.g., causal attention attribution) were not included.
\end{enumerate}

\subsection{Dataset-Specific vs.\ Merged Training}
\label{sec:disc_merged}

Our current study merges three dialogue datasets (DSTC9, Wizard of Wikipedia, FaithDial) to create a unified training corpus.
While this approach demonstrates that progressive training is effective across dialogue types, it may obscure important task-specific differences in citation learning dynamics:

\begin{itemize}[leftmargin=*, itemsep=2pt]
    \item \textbf{Task-oriented dialogue (DSTC9)} focuses on FAQ-style knowledge with specific, goal-directed citations.
    \item \textbf{Open-domain dialogue (WoW)} uses Wikipedia articles for exploratory, conversational citations.
    \item \textbf{Hallucination-aware dialogue (FaithDial)} emphasises cautious, verifiable citations.
\end{itemize}

Future work should investigate whether citation learning strategies differ across these dialogue types by training models separately on each dataset.
Additionally, separate English and Hindi training tracks (rather than bilingual mixed training) would isolate language-specific effects from cross-lingual transfer effects.


\section{Conclusion and Future Work}
\label{sec:conclusion}

We presented \ours{}, a progressive four-stage training pipeline for explainable, knowledge-grounded dialogue generation in a bilingual English--Hindi setting. Through a comprehensive empirical study across six models (250M--7B parameters, encoder-decoder and decoder-only architectures), we demonstrated several key findings:

\begin{enumerate}[leftmargin=*, itemsep=2pt]
    \item \textbf{Citation-grounded SFT substantially reduces hallucination.} Training with explicit citation format reduces hallucination to 0.0\% under automatic NLI-based evaluation for encoder-decoder models, and to 0.010--0.014 for Mistral-7B. Notably, LLaMA-3.2-1B also achieves 0.0\% English hallucination after Stage~2 - but without generating any citation markers - demonstrating that hallucination elimination and citation format learning are separable outcomes. Human evaluation remains future work.

    \item \textbf{Progressive training prevents catastrophic forgetting.} English metrics remain stable while Hindi capabilities improve substantially through Stages~1--3, validating the incremental skill-composition approach.

    \item \textbf{SFT as a strong baseline for structured grounded tasks.} In our experimental configuration, GRPO alignment provided marginal improvement over SFT. This suggests that well-designed SFT may be a competitive baseline for citation-grounded dialogue, though the role of RL under different configurations warrants further investigation.

    \item \textbf{Small models match large models after SFT.} A 250M-parameter model achieves identical English performance to a 780M model after Stage~2, with implications for cost-effective deployment.

    \item \textbf{Generation collapse is recoverable.} Flan-T5-XL (3B) exhibited generation collapse at Stage~2 but recovered fully at Stage~3, demonstrating the resilience of continued training and highlighting the need for generation-time validation.
\end{enumerate}

\subsection{Future Work}

Several directions emerge from this study:

\paragraph{Improved GRPO training.} Reducing the KL penalty, increasing training steps, and designing graded reward signals may unlock RL benefits.

\paragraph{Human evaluation.} Complementing automatic metrics with human judgments would provide a more complete evaluation picture.

\paragraph{Additional languages.} Extending the pipeline to other Indian languages (Bengali, Tamil, Marathi) would test generalizability.

\paragraph{Explainability-guided training.} Using attention visualization insights to design training objectives that encourage faithful attention over cited knowledge passages.

\medskip
This work establishes a progressive training methodology for citation-grounded dialogue and provides baseline results across six language models. Future work will examine dataset-specific effects by training separately on DSTC9, Wizard of Wikipedia, and FaithDial, as well as language-specific effects through separate English and Hindi training tracks.

\bibliography{references}

\onecolumn
\appendix

\setlength{\floatsep}{6pt}
\setlength{\textfloatsep}{6pt}
\setlength{\intextsep}{4pt}
\setlength{\abovecaptionskip}{4pt}
\setlength{\belowcaptionskip}{2pt}

\noindent\rule{\textwidth}{0.4pt}
\vspace{0.3em}
\begin{center}
{\Large\bfseries Appendix A: LLaMA-3.2-1B Generated Output Examples}
\end{center}
\vspace{0.3em}
\noindent\rule{\textwidth}{0.4pt}
\vspace{0.5em}

Table~\ref{tab:llama_outputs} presents representative Stage~3 outputs from LLaMA-3.2-1B alongside the expected citation-grounded reference. English outputs contain factually correct content but no citation markers; Hindi outputs correctly produce \texttt{[N]} markers integrated into natural-language phrasing.

\begin{table}[H]
\centering
\caption{Representative LLaMA-3.2-1B Stage~3 outputs vs.\ expected references. EN outputs contain zero citation markers despite the model having access to the numbered knowledge passage. HI outputs correctly produce \texttt{[N]} markers.}
\label{tab:llama_outputs}
\small
\begin{tabular}{@{}p{0.04\textwidth}p{0.19\textwidth}p{0.22\textwidth}p{0.32\textwidth}p{0.05\textwidth}@{}}
\toprule
\textbf{L} & \textbf{Query (excerpt)} & \textbf{Expected Output} & \textbf{LLaMA-3.2-1B Output} & \textbf{Cit.} \\
\midrule
EN & Can you bring your own liquor to Pizza Hut Cherry Hinton? &
As noted in [1], No, outside beverages are not allowed. &
No, you cannot bring your own liquor at Pizza Hut Cherry Hinton. Do you have any other questions? Would you like to make a reservation?$\ldots$ &
\textbf{0} \\
\addlinespace
EN & How long did California Love stay on Billboard Hot 100? &
Based on [1], yeah, stayed on the billboard hot 100's for two weeks straight! &
It was one of 2Pac's most popular singles, reaching number one on the Billboard Hot 100 for two weeks. It was a double A-side single$\ldots$ &
\textbf{0} \\
\addlinespace
EN & Restaurant booking confirmation &
According to [1], booking was successful. Ref:~\texttt{ENOALFAV}. &
Your reservation is all set. The reference number is \texttt{1ZQD0Z1A}.$\ldots$ &
\textbf{0} \\
\midrule
HI & [\textit{Query about wheelchair accessibility at Prezzo}] &
{[2]}~\hi{में कहा गया है कि हां, प्रेज़ो में व्हीलचेयर उपलब्ध है।} &
{[2]}~\hi{के आधार पर, हां, प्रेज़ो में व्हीलचेयर उपलब्ध है इसलिए उन्हें आपकी सहायता करने में सक्षम होना चाहिए।} &
\textbf{[2] \checkmark} \\
\addlinespace
HI & [\textit{Query about the color black}] &
{[1]}~\hi{में उल्लेख है कि काले रंग के बारे में, यह पूरी तरह से दिखाई देने वाले प्रकाश को अवशोषित करता है।} &
{[1]}~\hi{में कहा गया है कि मुझे नहीं पता था कि यह था, काले रंग के बारे में}\ldots &
\textbf{[1] \checkmark} \\
\bottomrule
\end{tabular}
\end{table}

\vspace{1em}
\noindent\rule{\textwidth}{0.4pt}
\vspace{0.3em}
\begin{center}
{\Large\bfseries Appendix B: Figures}
\end{center}
\vspace{0.3em}
\noindent\rule{\textwidth}{0.4pt}
\vspace{0.5em}

\begin{figure}[H]
\centering
\begin{minipage}[t]{0.48\textwidth}
    \centering
    \includegraphics[width=\textwidth]{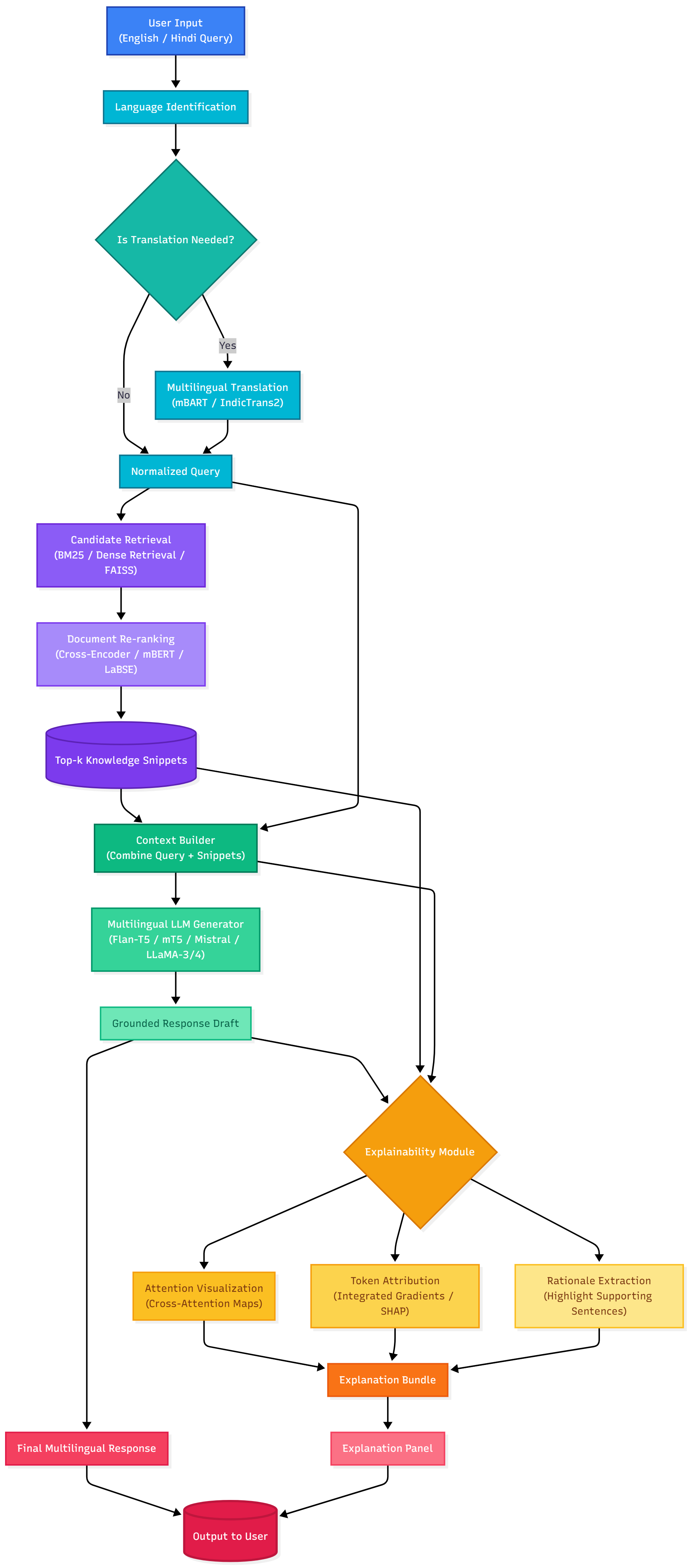}
    \captionof{figure}{System architecture of \ours{}. User queries undergo language identification and optional translation before retrieval. The context builder combines the query with top-$k$ knowledge snippets. The multilingual LLM generator produces a citation-grounded response, which is analyzed by the explainability module.}
    \label{fig:architecture}
\end{minipage}\hfill
\begin{minipage}[t]{0.48\textwidth}
    \centering
    \includegraphics[width=\textwidth]{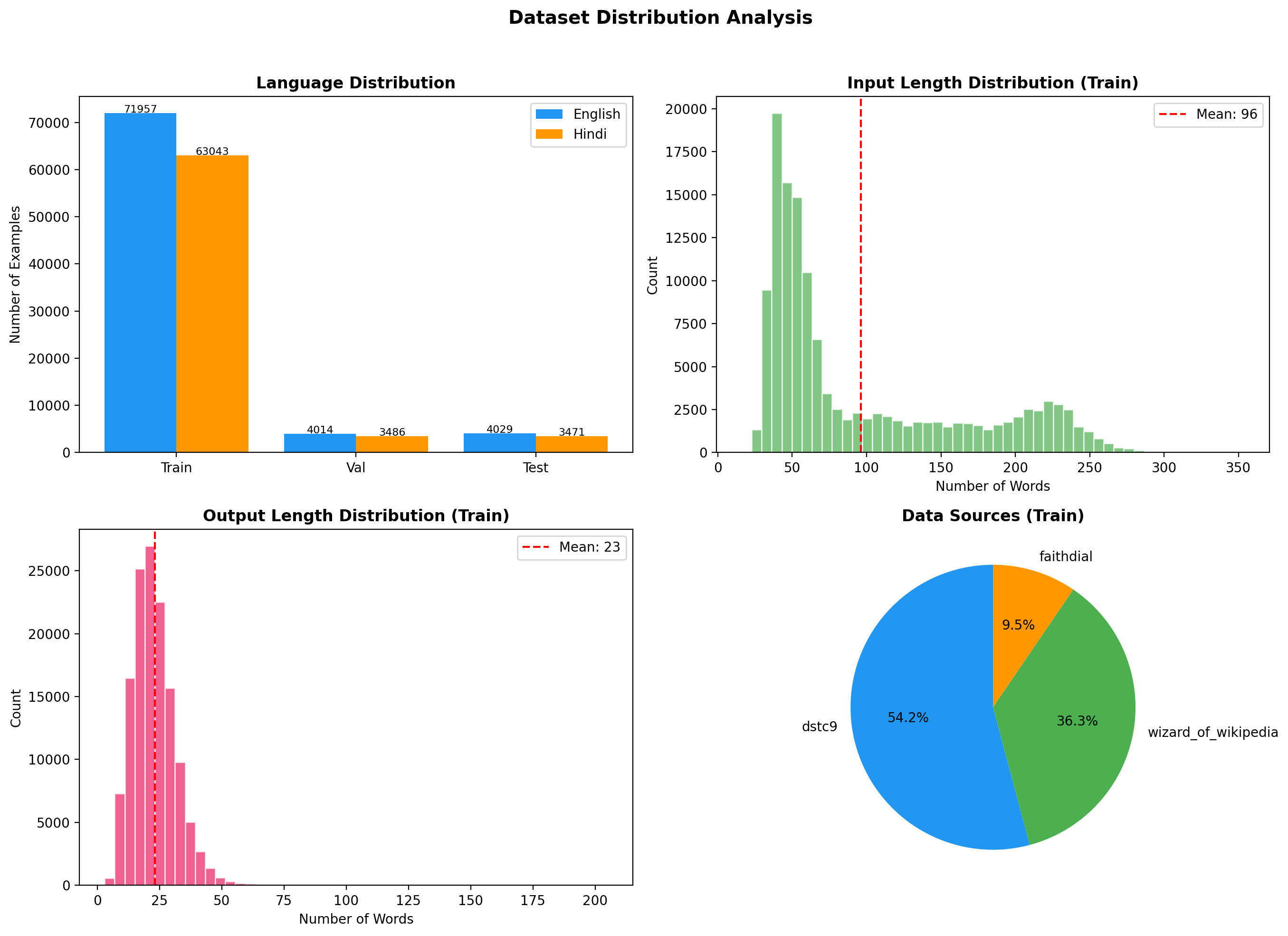}
    \captionof{figure}{Data distribution across source corpora (DSTC9, FaithDial, Wizard of Wikipedia) and language splits (English, Hindi) across the train, validation, and test partitions.}
    \label{fig:data_dist}
\end{minipage}
\end{figure}

\begin{figure}[H]
\centering
\begin{minipage}[t]{0.48\textwidth}
    \centering
    \includegraphics[width=\textwidth]{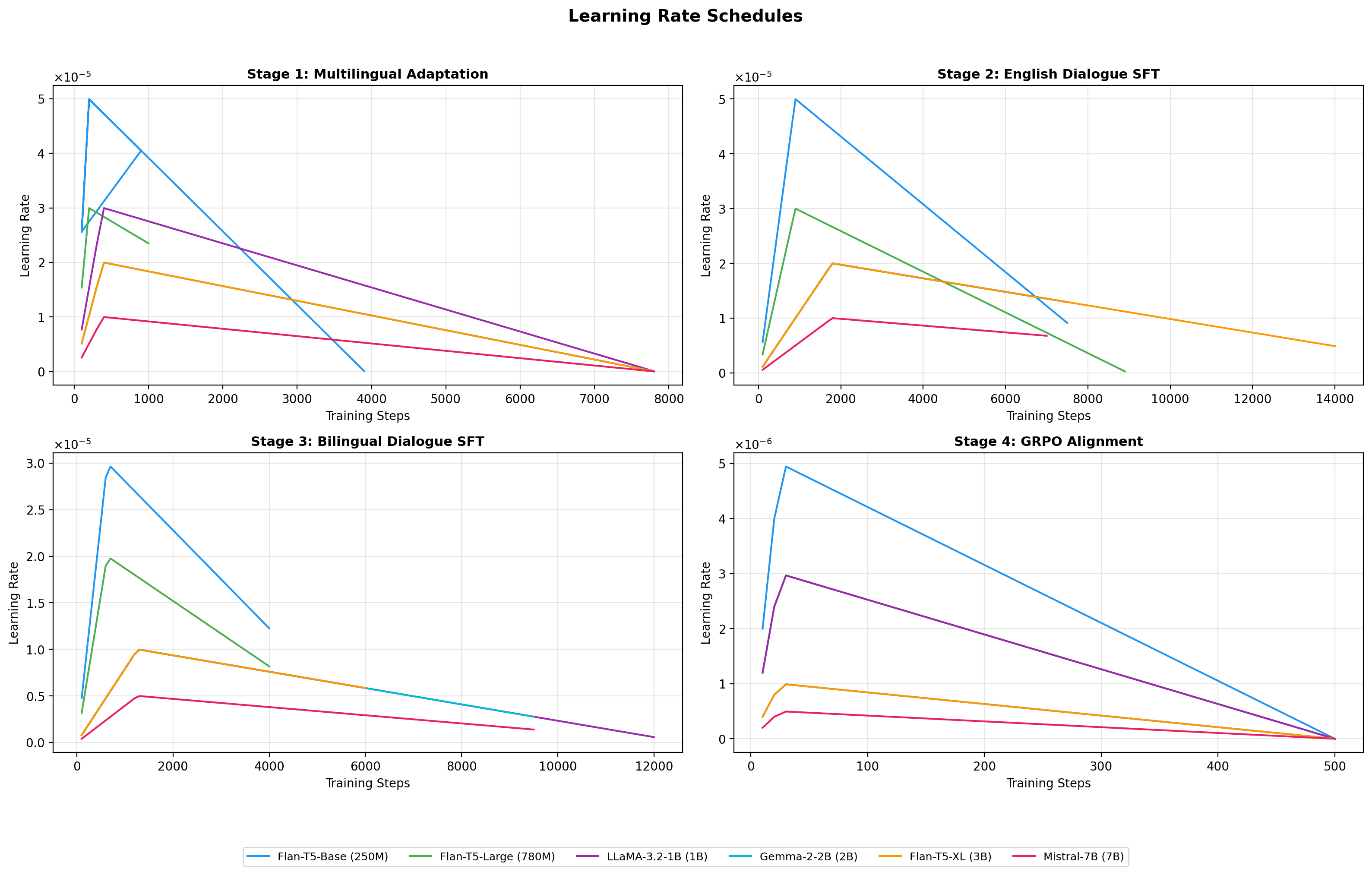}
    \captionof{figure}{Learning rate schedules across all training stages and models. Cosine decay with warmup is used for SFT stages; a lower learning rate with linear warmup is used for GRPO.}
    \label{fig:lr_schedules}
\end{minipage}\hfill
\begin{minipage}[t]{0.48\textwidth}
    \centering
    \includegraphics[width=\textwidth]{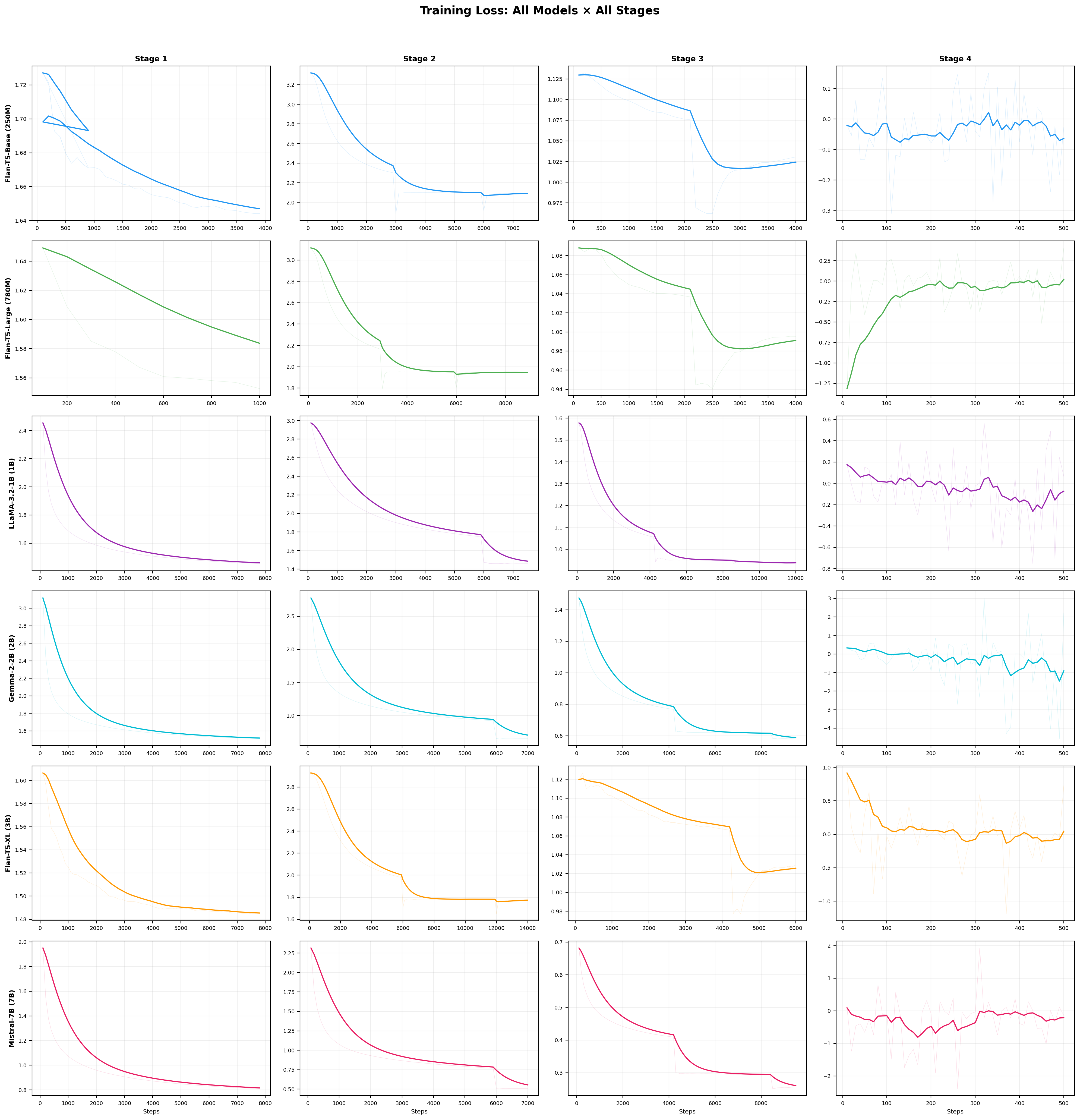}
    \captionof{figure}{Training loss progression across all stages for all six models. Stage~2 (English SFT) and Stage~3 (Bilingual SFT) show consistent convergence across architectures.}
    \label{fig:loss_grid}
\end{minipage}
\end{figure}

\begin{figure}[H]
\centering
\begin{minipage}[t]{0.48\textwidth}
    \centering
    \includegraphics[width=\textwidth]{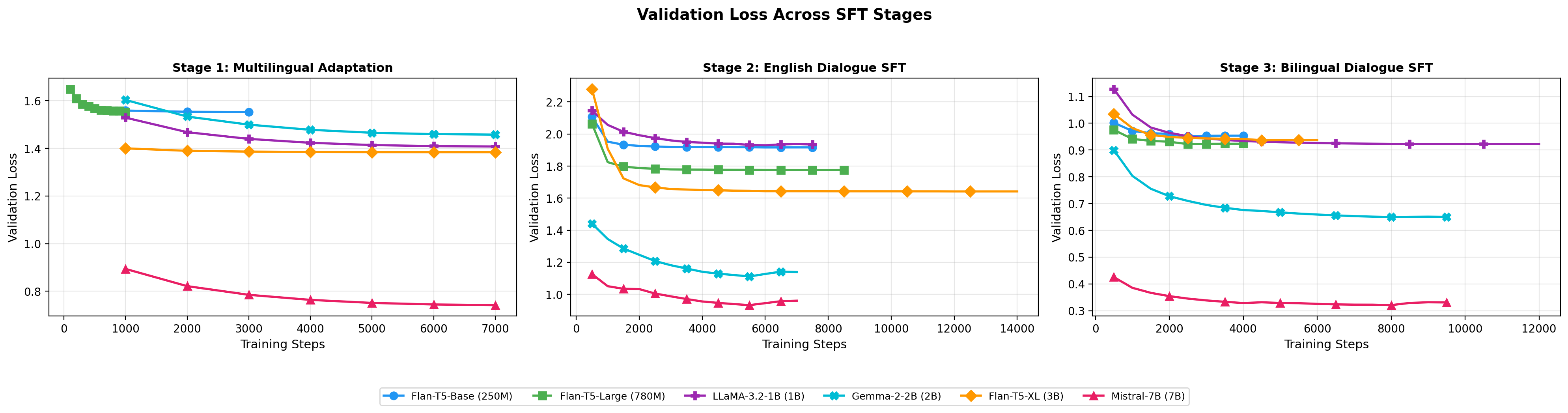}
    \captionof{figure}{Validation loss across SFT stages for all six models. Mistral-7B achieves the lowest loss throughout; Gemma-2-2B reaches 0.651 at Stage~3. Flan-T5-Large (dashed line, Stage~1) uses training loss as fallback since eval loss was not logged for that stage.}
    \label{fig:val_loss}
\end{minipage}\hfill
\begin{minipage}[t]{0.48\textwidth}
    \centering
    \includegraphics[width=\textwidth]{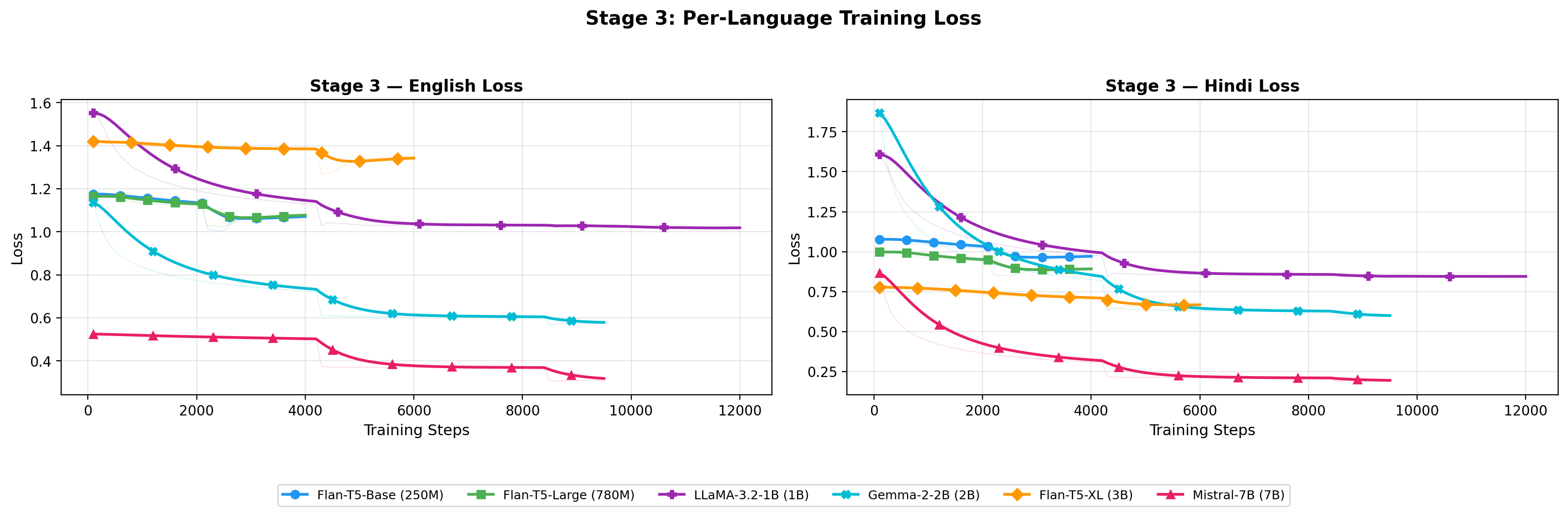}
    \captionof{figure}{Language-specific training loss during Stage~3 (Bilingual SFT). Hindi loss starts higher but decreases rapidly, converging with English.}
    \label{fig:lang_loss}
\end{minipage}
\end{figure}

\begin{figure}[H]
\centering
\begin{minipage}[t]{0.48\textwidth}
    \centering
    \includegraphics[width=\textwidth]{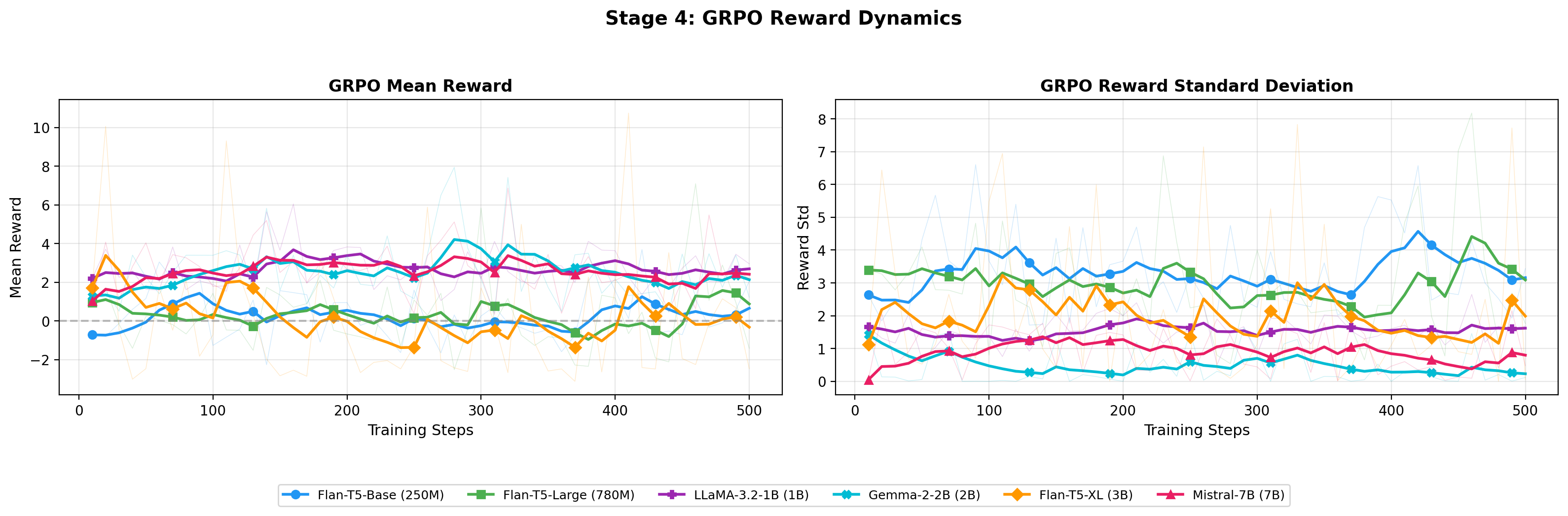}
    \captionof{figure}{GRPO reward trajectory during Stage~4 training for all six models. Gemma-2-2B achieves the highest absolute reward (3.40) with no decline; most other models show reward regression from best to final.}
    \label{fig:grpo_rewards}
\end{minipage}\hfill
\begin{minipage}[t]{0.48\textwidth}
    \centering
    \includegraphics[width=\textwidth]{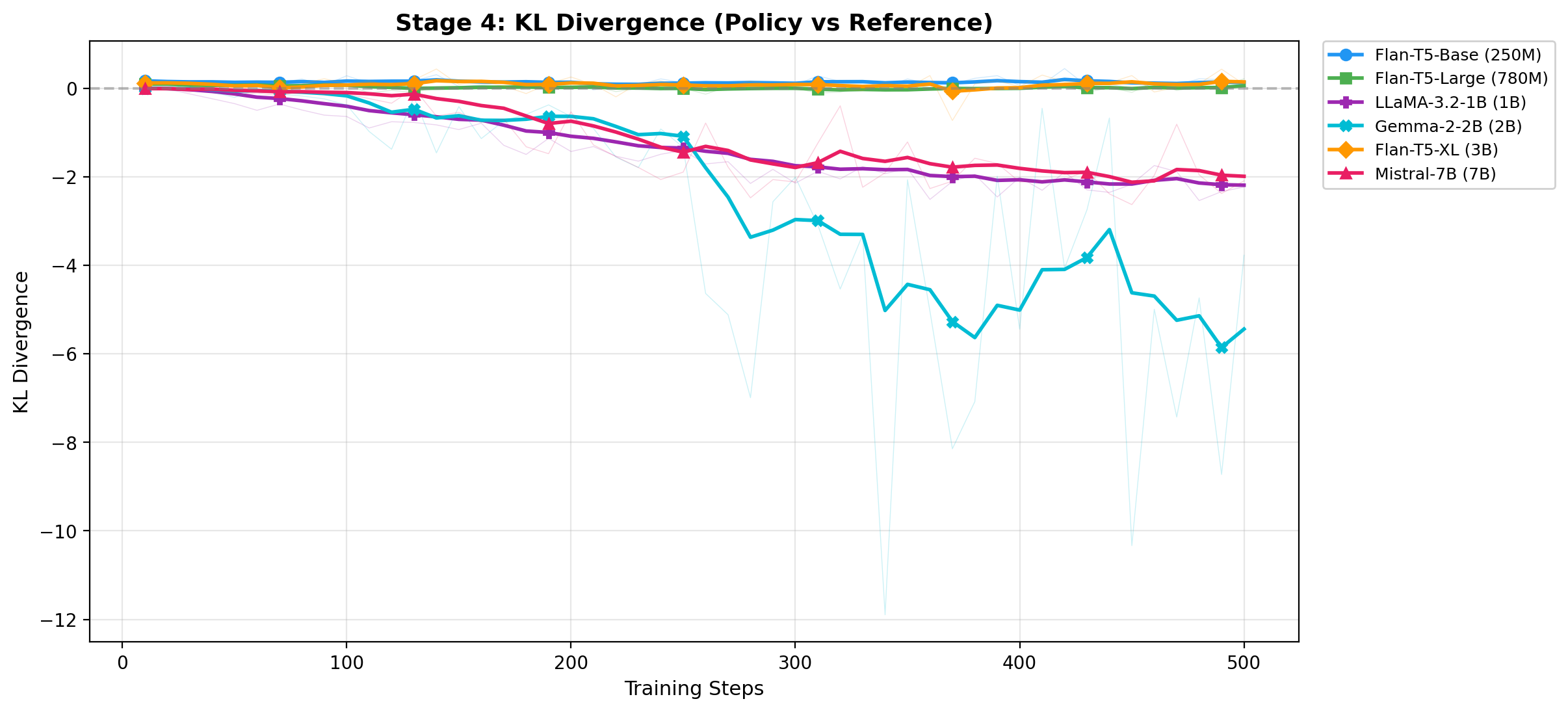}
    \captionof{figure}{KL divergence between the policy and reference model during GRPO training. The KL penalty ($\beta = 0.04$) constrains policy drift.}
    \label{fig:grpo_kl}
\end{minipage}
\end{figure}

\vspace{1em}
\noindent\rule{\textwidth}{0.4pt}
\vspace{0.3em}
\begin{center}
{\Large\bfseries Appendix C: Additional Evaluation Figures}
\end{center}
\vspace{0.3em}
\noindent\rule{\textwidth}{0.4pt}
\vspace{0.5em}

\begin{figure}[H]
\centering
\begin{minipage}[t]{0.48\textwidth}
    \centering
    \includegraphics[width=\textwidth]{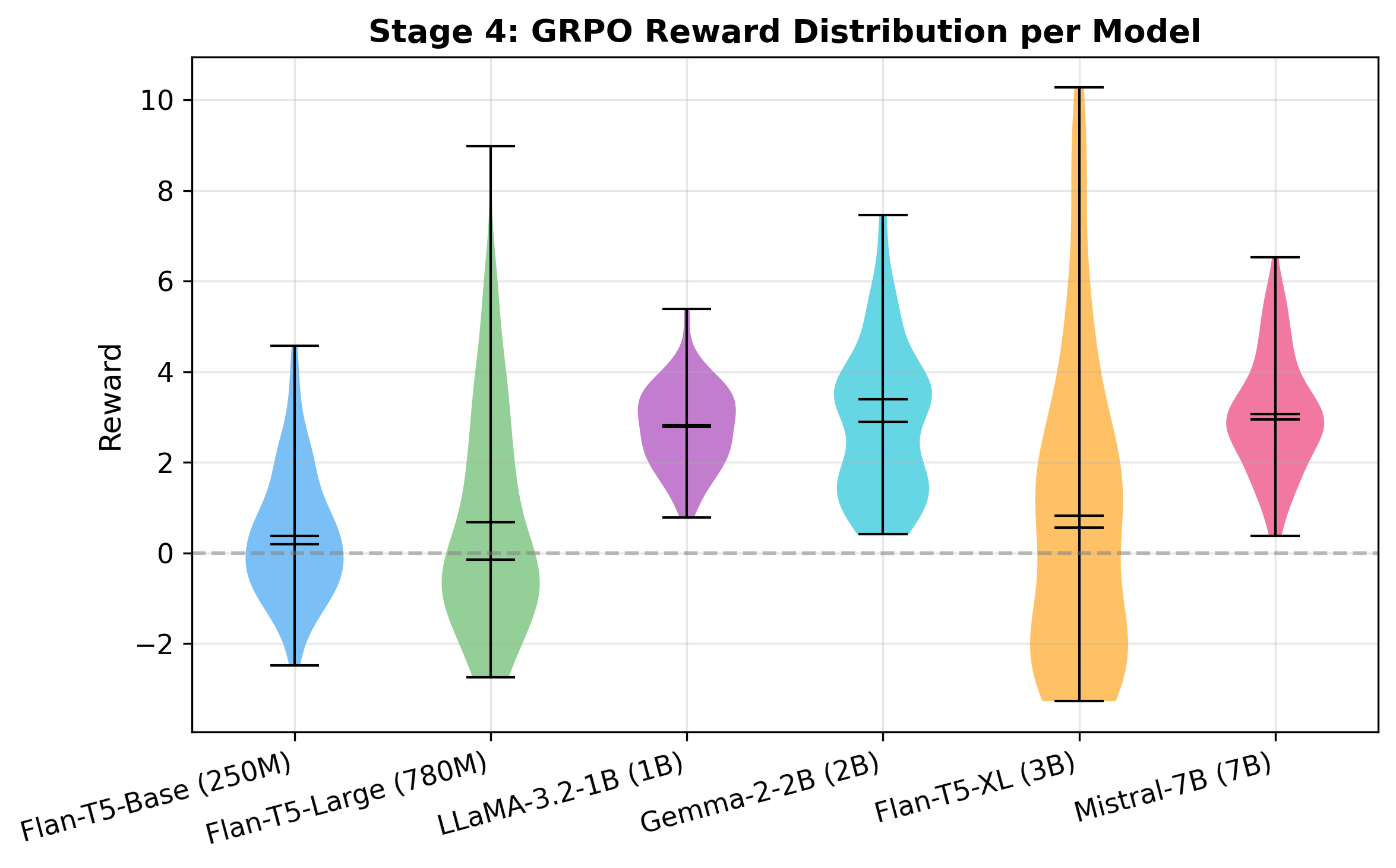}
    \captionof{figure}{Violin plot of GRPO reward distributions per model. Mistral-7B (7B) shows a compact, high-reward distribution centered around 3.0, while encoder-decoder models show wider spread with lower medians.}
    \label{fig:grpo_reward_distribution}
\end{minipage}\hfill
\begin{minipage}[t]{0.48\textwidth}
    \centering
    \includegraphics[width=\textwidth]{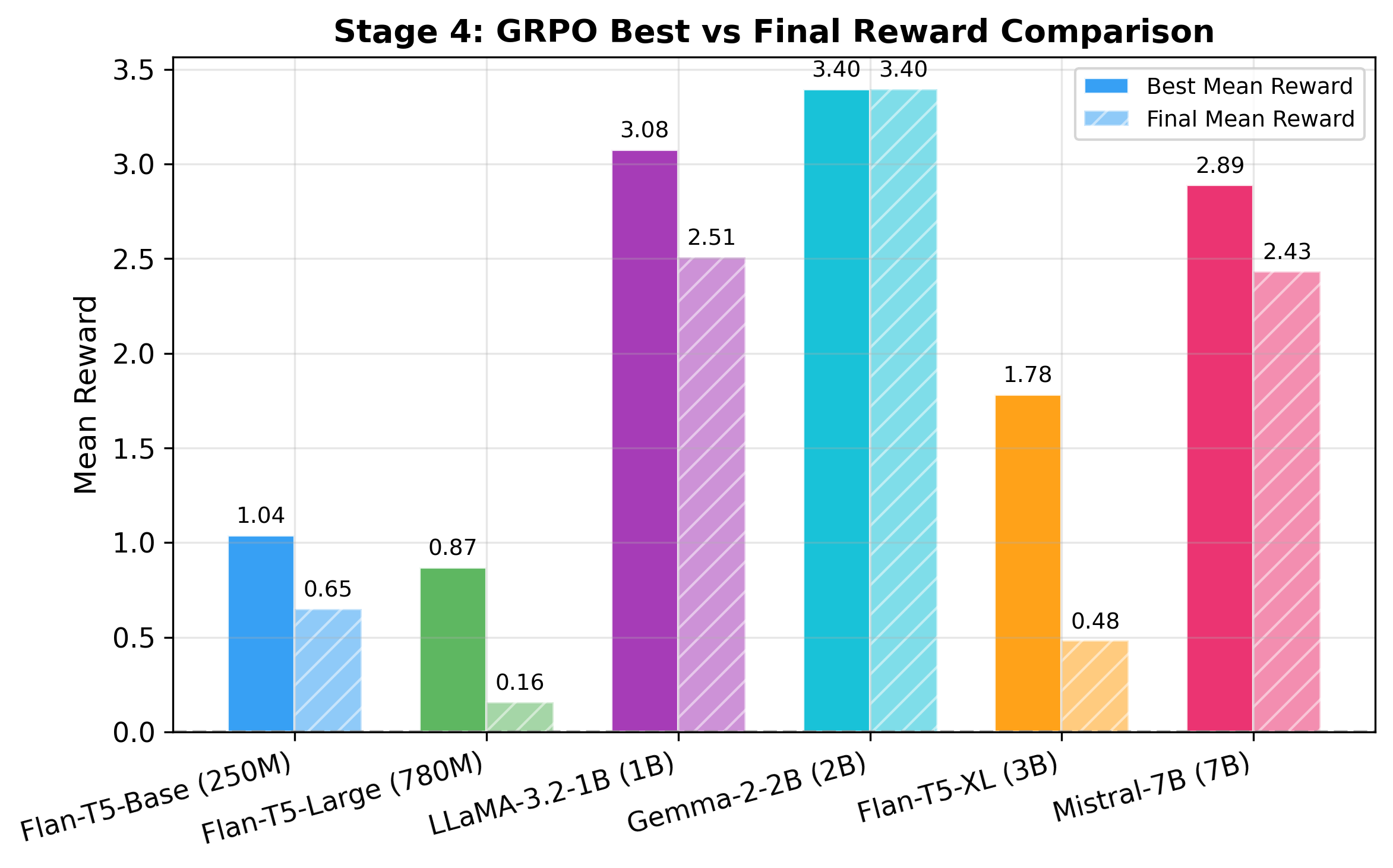}
    \captionof{figure}{GRPO best vs.\ final mean reward comparison. All models show reward decline from best to final, with Flan-T5-XL exhibiting the largest drop ($-$1.30) and Mistral-7B the smallest ($-$0.46).}
    \label{fig:grpo_reward_comparison}
\end{minipage}
\end{figure}

\begin{figure}[H]
\centering
\includegraphics[width=0.6\textwidth]{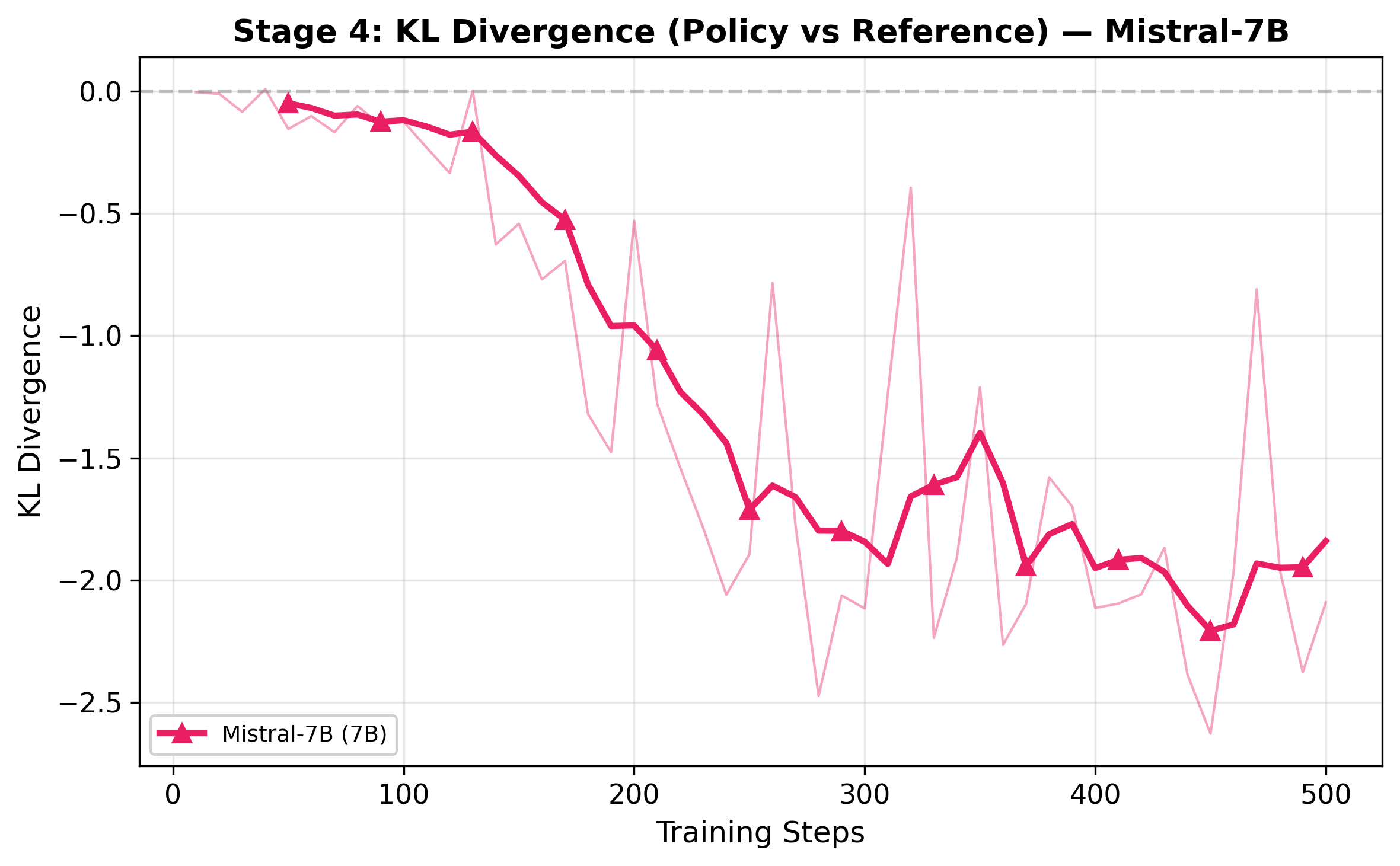}
\caption{KL divergence (policy vs.\ reference) for Mistral-7B during GRPO. The negative KL indicates the policy diverges from the reference, stabilizing around $-$2.0 after 300 steps despite the KL penalty ($\beta = 0.04$).}
\label{fig:grpo_kl_mistral}
\end{figure}

\end{document}